\documentclass[dvipsnames]{article} 
\usepackage{nips15submit_e,times}
\usepackage{hyperref}
\usepackage{url}

\usepackage{subfigure}
\usepackage{xcolor}
\usepackage{tikz}
\usetikzlibrary{bayesnet, decorations.pathreplacing}

\usepackage{dcolumn}

\usepackage{amsmath}
\usepackage{amssymb}
\usepackage{amsbsy}
\usepackage{url}
\usepackage{enumitem}

\newcommand{\bx}{\mathbf{x}}
\newcommand{\bR}{\mathbf{R}}
\newcommand{\tbx}{\tilde{\mathbf{x}}}

\newcommand{\bu}{\mathbf{u}}
\newcommand{\bz}{\mathbf{z}}
\newcommand{\hbz}{\hat{\mathbf{z}}}

\newcommand{\bs}{\mathbf{s}}
\newcommand{\bv}{\mathbf{v}}
\newcommand{\br}{\mathbf{r}}
\newcommand{\bp}{\mathbf{p}}
\newcommand{\bW}{\mathbf{W}}
\newcommand{\bb}{\mathbf{b}}
\newcommand{\bA}{\mathbf{A}}
\newcommand{\bB}{\mathbf{B}}
\newcommand{\bo}{\mathbf{o}}

\newcommand{\bC}{\mathbf{C}}
\newcommand{\bH}{\mathbf{H}}
\newcommand{\bI}{\mathbf{I}}

\newcommand{\tlat}{\text{lat}}

\newcommand{\mN}{\mathcal{N}}
\newcommand{\bSigma}{\mathbf{\Sigma}}
\newcommand{\bsigma}{{\boldsymbol{\sigma}}}
\newcommand{\bmu}{{\boldsymbol{\mu}}}
\newcommand{\zgoal}{\bz_{\text{goal}}}
\newcommand{\cL}{\mathcal{L}}

\renewcommand\b[1]{{\ensuremath\boldsymbol{#1}}}

\newcommand{\bphi}{{\boldsymbol{\phi}}}
\newcommand{\btheta}{{\boldsymbol{\theta}}}
\newcommand{\bpsi}{{\boldsymbol{\psi}}}

\newcommand{\dec}{h^{\text{dec}}_\btheta}
\newcommand{\enc}{h^{\text{enc}}_\bphi}
\newcommand{\trans}{h^{\text{trans}}_\bpsi}

\usepackage{float}

\definecolor{dgreen}{rgb}{0,.7,0}
\definecolor{dyellow}{rgb}{.7,.7,0}
\definecolor{dred}{rgb}{.7,0,0}
\definecolor{dblue}{rgb}{0,0,0.7}

\newcommand{\hide}[1]{}

\usepackage[numbers,square]{natbib}

\DeclareMathOperator{\Tr}{Tr}

\title{Embed to Control: A Locally Linear Latent Dynamics Model for Control from Raw Images}

\author{
Manuel Watter\thanks{Authors contributed equally.}~ ~~~~~~~~~ Jost Tobias Springenberg\footnotemark[1] \\ \textbf{Joschka Boedecker} \\
University of Freiburg, Germany\\
\small{\texttt{\{watterm,springj,jboedeck\}@cs.uni-freiburg.de}}
\And
Martin Riedmiller \\
Google DeepMind \\
London, UK \\ 
\small{\texttt{riedmiller@google.com}}
}

\nipsfinalcopy 

\begin{document}

\maketitle

\begin{abstract}
We introduce Embed to Control (E2C), a method for model learning and control of non-linear dynamical systems from raw pixel images. E2C consists of a deep generative model, belonging to the family of variational autoencoders, that learns to generate image trajectories from a latent space in which the dynamics is constrained to be locally linear. Our model is derived directly from an optimal control formulation in latent space, supports long-term prediction of image sequences and exhibits strong performance on a variety of complex control problems. 
\end{abstract}

\section{Introduction}
\label{sect:intro}
Control of non-linear dynamical systems with continuous state and action spaces is one of
the key problems in robotics and, in a broader context, in reinforcement learning for autonomous agents. 
A prominent class of algorithms that aim to solve this problem are model-based locally optimal (stochastic) control algorithms such as iLQG control \cite{Jacobson1970,Todorov2005}, which approximate the general non-linear control problem via local linearization. When combined with receding horizon control \cite{Tassa_2008}, and ma\-chine learning methods for learning approximate system models, such algorithms are powerful tools for solving complicated control problems \cite{Tassa_2008,Pan_2014,Levine_2013n};
however, they either rely on a known system model or require the design of relatively low-dimensional state representations. 
 For real \emph{autonomous} agents to succeed, we ultimately need
algorithms that are capable of controlling complex dynamical systems
from \emph{raw sensory input (e.g. images) only}.
In this paper we tackle this difficult problem.

If stochastic optimal control (SOC) methods were applied directly to control from raw image data, they would face two major obstacles. First, sensory data is usually high-dimensional -- i.e. images with thousands of pixels -- rendering a naive SOC solution computationally infeasible. Second, the image content is typically a highly non-linear function of the system dynamics underlying the observations; thus model identification and control of this dynamics are non-trivial.

While both problems could, in principle, be addressed by designing more advanced SOC algorithms 
we approach the ``optimal control from raw images'' problem differently: 
turning the problem of locally optimal control in high-dimensional non-linear systems into one of identifying a low-dimensional latent state space, in which locally optimal control can be performed robustly and easily. To learn such a latent space we propose a new deep generative model belonging to the class of variational autoencoders \cite{Kingma_ICLR2014,Rezende_ICML2014} that is derived from an iLQG formulation in latent space. The resulting \emph{Embed to Control (E2C)} system is a probabilistic generative model that holds a belief over viable trajectories in sensory space, allows for accurate long-term planning in latent space, and is trained fully unsupervised.
We demonstrate the success of our approach on four challenging tasks for control from raw images
and compare it to a range of methods for unsupervised representation
learning. As an aside, we also validate that deep up-convolutional
networks \cite{Zeiler_CVPR2010,DosSprChair15} are powerful generative models for large images.

\section{The Embed to Control (E2C) model}
\label{sect:method}
We briefly review the problem of SOC for dynamical systems, introduce approximate locally optimal control in latent space, and finish with the derivation of our model.

\subsection{Problem Formulation}
We consider the control of unknown dynamical systems of the form
\begin{equation}
  \bs_{t+1} = f(\bs_t, \bu_t) + \b \xi, \ \ \b \xi \sim \mN(0, \bSigma_{\b \xi}),
\label{eq:dynamics}
\end{equation}
where $t$ denotes the time steps, $\bs_t \in \mathbb{R}^{n_s}$ the system state, 
$\bu_t \in \mathbb{R}^{n_u}$ the applied control and $\b \xi$ the system noise. The function $f(\bs_t, \bu_t)$ is an arbitrary, smooth, system dynamics. 
We equivalently refer to Equation \eqref{eq:dynamics} using the notation $P(\bs_{t+1} | \bs_t, \bu_t)$, which we assume to be a multivariate normal distribution $\mN(f(\bs_t, \bu_t),  \bSigma_\xi)$.
We further assume that 
we are only  given access to visual depictions $\bx_t \in \mathbb{R}^{n_x}$ of state $\bs_t$. This restriction 
requires solving a joint state identification and control problem.
For simplicity we will in the following assume that $\bx_t$ is a fully observed depiction of $\bs_t$, but relax this assumption later. 

Our goal then is to infer a low-dimensional latent state space model in which optimal control can be performed. That is, we seek to learn a function $m$, mapping from high-dimensional images $\bx_t$ to low-dimensional vectors $\bz_t \in \mathbb{R}^{n_z}$ with $n_z \ll n_x$, such that the control problem can be solved using $\bz_t$ instead of $\bx_t$:
\begin{equation}
  \bz_t = m(\bx_t) + \b \omega, \ \ \b \omega \sim \mN(0, \bSigma_{\b \omega}),
\label{eq:gaussian_z} 
\end{equation}
where $\b \omega$ accounts for system noise; or equivalently $\bz_t \sim \mN(m(\bx_t), \bSigma_{\b \omega})$. 
Assuming for the moment that such a function can be learned (or approximated), we will first define SOC in a latent space and introduce our model thereafter.

\subsection{Stochastic locally optimal control in latent spaces}
\label{sect:soc}
Let $\bz_t \in \mathbb{R}^{n_z}$ be the inferred latent state from image $\bx_t$ of state $\bs_t$ and $f^{\tlat}(\bz_t, \bu_t)$ the transition dynamics in latent space, i.e., $\bz_{t+1} = f^{\tlat}(\bz_t, \bu_t)$. Thus $f^{\tlat}$ models the changes that occur in $\bz_t$ when control $\bu_t$ is applied to the underlying system as a latent space analogue to $f(\bs_t, \bu_t)$. Assuming $f^{\tlat}$ is known, optimal controls for a trajectory of length $T$ in the dynamical system can be derived by minimizing the function $J(\bz_{1:T}, \bu_{1:T})$ which gives the expected future costs when following $(\bz_{1:T}, \bu_{1:T})$:
\begin{equation}
  J(\bz_{1:T}, \bu_{1:T}) = \mathbb{E}_{\bz} \left [ c_T(\bz_T, \bu_T) + \sum_{t_0}^{T-1} c(\bz_t, \bu_t) \right ],
\end{equation}
where $c(\bz_t, \bu_t)$ are instantaneous costs, $c_T(\bz_T, \bu_T)$ denotes terminal costs and $\bz_{1:T} = \lbrace \bz_1, \dots, \bz_T \rbrace$ and $\bu_{1:T} = \lbrace \bu_1, \dots, \bu_T \rbrace$ are state and action sequences respectively. If $\bz_t$ contains sufficient information about $\bs_t$, i.e., $\bs_t$ can be inferred from $\bz_t$ alone, and $f^{\tlat}$ is differentiable, the cost-minimizing controls can be computed from $J(\bz_{1:T}, \bu_{1:T})$ via SOC algorithms \cite{Stengel_Book1994}. These optimal control algorithms approximate the global non-linear dynamics with locally linear dynamics at each time step $t$. Locally optimal actions can then be found in closed form. 
Formally, given a reference trajectory $\bar{\bz}_{1:T}$ -- the current estimate for the optimal trajectory -- together with corresponding controls $\bar{\bu}_{1:T}$ the system is linearized as
\begin{equation}
  \bz_{t+1} = \bA (\bar{\bz}_t) \bz_t + \bB (\bar{\bz}_t) \bu_{t} + \bo (\bar{\bz}_t) + \b \omega, \ \ \b \omega \sim \mN(0, \bSigma_{\b \omega}),
\label{eq:dynamics_latent}
\end{equation}
where $\bA (\bar{\bz}_t) = \frac{\delta f^{\tlat}(\bar{\bz}_t, \bar{\bu}_t)}{\delta \bar{\bz}_t}$, $\bB (\bar{\bz}_t) = \frac{\delta f^{\tlat}(\bar{\bz}_t, \bar{\bu}_t)}{\delta \bar{\bu}_t}$ are local Jacobians, and $\bo(\bar{\bz}_t)$ is an offset.
To enable efficient computation of the local controls we assume the costs to be a quadratic function of the latent representation
\begin{equation}
  c(\bz_t, \bu_t) = (\bz_t - \zgoal)^T \mathbf{R}_{z} (\bz_t - \zgoal) + \bu^T_t \mathbf{R}_{u} \bu_t,
  \label{eq:costs}
\end{equation}
where $\mathbf{R}_{z} \in \mathbb{R}^{n_z \times n_z}$ and $\mathbf{R}_{u} \in \mathbb{R}^{n_u \times n_u}$ are cost weighting matrices and $\zgoal$ is the inferred representation of the goal state. We also assume $c_T(\bz_T, \bu_T) = c(\bz_T, \bu_T)$ throughout this paper. In combination with Equation \eqref{eq:dynamics_latent} this gives us a local \emph{linear-quadratic-Gaussian} formulation at each time step $t$ which can be solved by SOC algorithms such as iterative linear-quadratic regulation (iLQR) \cite{Li_ICINC2004} or approximate inference control (AICO) \cite{Toussaint_ICML2009}. The result of this trajectory optimization step is a locally optimal trajectory with corresponding control sequence $(\bz^*_{1:T}, \bu^*_{1:T}) \approx \arg \min_{\substack{\bz_{1:T} \\ \bu_{1:T}}} J(\bz_{1:T}, \bu_{1:T})$.

\subsection{A locally linear latent state space model for dynamical systems}
\label{sect:model}
Starting from the SOC formulation, we now turn to the problem of learning an appropriate low-dimensional latent representation $\bz_t \sim P(Z_t | m(\bx_t), \bSigma_{\b \omega})$ of $\bx_t$. 
The representation $\bz_t$ has to fulfill three properties: (i) it must capture sufficient information about $\bx_t$ (enough to enable reconstruction); 
(ii) it must allow for accurate prediction of the next latent state $\bz_{t+1}$ and thus, implicitly, of the next observation $\bx_{t+1}$; (iii) the prediction $f^\tlat$ of the next latent state must be locally linearizable \emph{for all valid control magnitudes} $\bu_t$.
Given some representation $\bz_t$, 
properties (ii) and (iii) in particular require us to capture possibly highly non-linear changes of the latent representation due to transformations of the observed scene induced by control commands. 
Crucially, these are particularly hard to model and subsequently linearize.
We circumvent this problem by taking a more direct approach: instead of learning a latent space $\bz$ and transition model $f^{\tlat}$ which are then linearized and combined with SOC algorithms, we directly impose desired transformation properties on the representation $\bz_t$ during learning. We will select these properties such that prediction in the latent space as well as locally linear inference of the next observation according to Equation \eqref{eq:dynamics_latent} are easy. 

The transformation properties that we desire from a latent
representation can be formalized directly from the iLQG formulation
given in Section \ref{sect:soc}~. 
Formally, following Equation \eqref{eq:gaussian_z}, let the latent
representation be Gaussian $P(Z|X) = \mathcal{N}(m(\bx_t), \bSigma_{\b \omega})$.
To infer $\bz_t$ from $\bx_t$ we first require a method for sampling latent states. 
Ideally, we would generate samples directly from the unknown true posterior $P(Z | X)$, which we, however, have no access to.
 Following the variational Bayes approach (see \citet{Jordan1999} for an overview) we resort to sampling $\bz_t$ from an approximate posterior distribution $Q_{\bphi}(Z | X)$ with parameters $\bphi$.

\begin{figure}[t]
	\centering
	\begin{tikzpicture}[outer sep=0]

\pgfdeclarelayer{background}
\pgfdeclarelayer{foreground}
\pgfsetlayers{background,main,foreground}

\def \mainxdist {0.9}
\def \mainydist {0.6}

\tikzstyle{box} = [rectangle, draw=black, rounded corners, minimum height=0.6cm]
\tikzstyle{close} = [inner sep=0pt, outer sep=0, line width=0.75mm]
\tikzstyle{con} = [shorten >=3pt, shorten <=3pt, >=latex, line width=0.75mm]
\tikzstyle{inter} = [thick, shorten <=2pt]
\tikzstyle{mlp} = [trapezium, trapezium angle=77,draw=black, shape border rotate=270]
\tikzstyle{obs} = [fill=black!30]
\tikzstyle{enc} = [draw=MidnightBlue!100, densely dotted]
\tikzstyle{dec} = [draw=Orange!100, dashed]
\tikzstyle{tr} = [draw=OliveGreen!100, solid]

\node [box, obs] (x_t) {$\mathbf{x}_t$};

\node [above right=0.1 * \mainydist and \mainxdist of x_t, mlp] (enc_left) {$h^{\text{enc}}_\phi$}
	(x_t.north) edge[con, enc, ->, in=180, out=75] (enc_left.west)
;

\node [below right=0.1 * \mainydist and 1.5 * \mainxdist of x_t, mlp] (dec_left) {$h^{\text{dec}}_\theta$}
;

\node [left=0.1 of dec_left, close] (p_t) {$\mathbf{p}_t$}
	(x_t.south) edge[con, dec, <-, in=180, out=315] (p_t.west)
;

\node [above right=-0.25 and 0.2 of enc_left, close] (mu_t) {$\mathbf{\mu}_t$}
;
\node [below right=-0.25 and 0.2 of enc_left, close] (sigma_t) {$\mathbf{\Sigma}_t$}
;

\node [above right=-0.1 and 0.3 of sigma_t] (gauss_t) {}
	(mu_t.east) edge[inter, close, enc, out=315, in=180] (gauss_t.center)
	(sigma_t.east) edge[inter, enc, close, in=180] (gauss_t.center)
;

\node [right= 3.5 * \mainxdist of x_t, box] (z_t) {$\mathbf{z}_t$}
	(gauss_t.center) edge[con, enc, ->, shorten <=0, in=145, out=0] (z_t.north)
	(dec_left.east) edge[con, dec, <-, in=245, out=0] (z_t.south)
;

\node [above right=0.2 * \mainydist and 0.5 * \mainxdist of z_t, mlp] (trans) {$h^{\text{trans}}_\psi$}
	(z_t.north) edge[con, tr, ->, in=215, shorten <=4] (trans.west)
;

\node [right=0.1 of trans, close] (B_t) {$\mathbf{B}_t$}
;
\node [above=0.1 of B_t, close] (A_t) {$\mathbf{A}_t$}
;
\node [below=0.1 of B_t, close] (b_t) {$\mathbf{o}_t$}
;

\node[below right=0.2 of z_t, box, obs] (a_t) {$\mathbf{u}_t$};

\node [box, right=2.5 * \mainxdist of z_t] (z_hat) {$\hat{\mathbf{z}}_{t+1}$}
	(A_t.east) edge[con, close, tr, ->, in=135, out=300] (z_hat.north) {}
	(B_t.east) edge[con, close, tr, ->, in=135, out=330] (z_hat.north) {}
	(b_t.east) edge[con, close, tr, ->, in=135, out=20] (z_hat.north) {}
	(z_t) edge[con, tr, ->] (z_hat.west) {}
	(a_t) edge[con, tr, ->, out=0, in=220] (z_hat.west) {}
;

\node [right=0.01cm of z_hat] (approx) {$\mathbf{\approx}$};

\node [box, right=0.01cm of approx] (z_t1) {$\mathbf{z}_{t+1}$};

\node [above=0.5 of z_hat.east, fill=white, xshift=-0.1cm] (q_hat) {\scriptsize $\hat{Q}_\bpsi$};
\node [above=0.52 of z_t1.west, fill=white, xshift=0.1cm] (q) {\scriptsize $Q_\bphi$};

\draw[decorate, decoration={brace, amplitude=10pt, raise=3pt}] (z_hat.north) -- (z_t1.north) node[black, midway, yshift=0.8cm, fill=white, inner sep=0] {\footnotesize $\text{KL}$};

\node [above right=0.1 * \mainydist and 1.5 * \mainxdist of z_t1, mlp, shape border rotate=90] (enc_right) {$h^{\text{enc}}_\phi$}
;

\node [above left=-0.25 and 0.1 of enc_right, close] (mu_t1) {$\mathbf{\mu}_{t+1}$}
;
\node [below left=-0.25 and 0.1 of enc_right, close] (sigma_t1) {$\mathbf{\Sigma}_{t+1}$}
;

\node [above left=-0.1 and 0.3 of sigma_t1] (gauss_t1) {}
	(mu_t1.west) edge[inter, close, enc, out=180, in=0] (gauss_t1.center)
	(sigma_t1.west) edge[inter, close, enc, out=180, in=0] (gauss_t1.center)
	(z_t1.north) edge[con, <-, enc, shorten >=0, in=210, out=30] (gauss_t1.center)
;

\node [below right=0.1 * \mainydist and 1 * \mainxdist of z_t1, mlp, shape border rotate=90] (dec_right) {$h^{\text{dec}}_\theta$}
	(z_hat.south) edge[con, dec, ->, in=180, out=320] (dec_right.west)
;

\node [right=0.1 of dec_right, close] (p_t1) {$\mathbf{p}_t$}
;

\node [box, obs, right=3.5 * \mainxdist of z_t1] (x_t1) {$\mathbf{x}_{t+1}$}
	(p_t1.east) edge[con, ->, dec, out=0, in=230] (x_t1.south)
	(enc_right.east) edge[con, <-, enc, out=0, in=130] (x_t1.north)
;

\begin{pgfonlayer}{background}
    \path[draw=black!30, dashed]
	   	([shift={(z_t.east |- enc_left.north)}] 0.05, 0.2) 
    		--
        ([shift={(z_t.east |- dec_left.south)}] 0.05, -0.2);
    \path[draw=black!30, dashed]
	   	([shift={(z_t1.west |- enc_left.north)}] -0.05, 0.2) 
    		--
        ([shift={(z_t1.west |- dec_left.south)}] -0.05, -0.2);
        
	\draw [con, enc, ->, line width=0.4mm] (dec_right.east) ++(0.7, -0.3) -- +(1cm,0) node [draw=none, inner sep=0, shift={(0.2, 0.03)}] {\tiny \textcolor{MidnightBlue}{encode}};
	\draw [con, dec, ->, line width=0.4mm] (dec_right.east) ++(0.7, -0.5) -- +(1cm,0) node [draw=none, inner sep=0, shift={(0.2, 0.03)}] {\tiny \textcolor{Orange}{decode}};
	\draw [con, tr, ->, line width=0.4mm] (dec_right.east) ++(0.7, -0.7) -- +(1cm,0) node [draw=none, inner sep=0, shift={(0.3, 0.03)}] {\tiny \textcolor{OliveGreen}{transition}};
\end{pgfonlayer}

\end{tikzpicture}
    \vspace{-0.2cm}
    \caption{The information flow in the E2C model. From left to right, we encode and decode an image $\bx_t$ with the networks $\enc$ and $\dec$, where we use the latent code $\bz_t$ for the transition step. The $\trans$ network computes the local matrices $\bA_t, \bB_t, \bo_t$ with which we can predict $\mathbf{\hat{z}}_{t+1}$ from $\bz_t$ and $\bu_t$. Similarity to the encoding $\bz_{t+1}$ is enforced by a KL divergence on their distributions and reconstruction is again performed by $\dec$.}
    \label{fig:model}
\vspace{-0.3cm}
\end{figure}
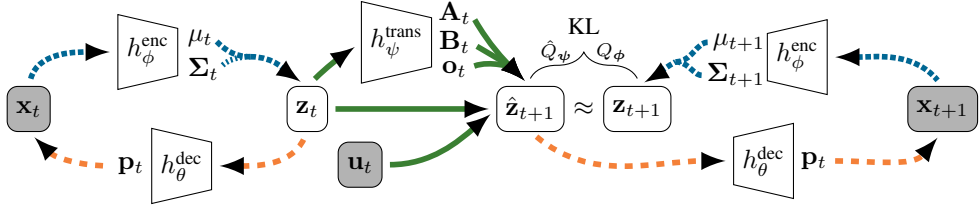

\textbf{Inference model for $Q_\bphi$.}
In our work this is always a diagonal Gaussian distribution $Q_\bphi(Z | X) = \mN(\bmu_t, \text{diag}(\bsigma^2_t))$, whose mean $\bmu_t \in \mathbb{R}^{n_z}$ and covariance $\bSigma_t = \text{diag}(\bsigma^2_t) \in \mathbb{R}^{n_z \times n_z}$ are computed by an encoding neural network with outputs
\begin{align}
\bmu_t &= \bW_\bmu \enc(\bx_t)  + \bb_{\b\mu},  \\
\log \b\sigma_t &= \bW_\bsigma \enc(\bx_t) + \bb_{\b\sigma},
\end{align}
where $\enc \in \mathbb{R}^{n_e}$ is the activation of the last hidden layer and where $\bphi$ is given by the set of all learnable parameters of the encoding network, including the weight matrices $\bW_\bmu$, $\bW_\bsigma$ and biases $\bb_\bmu$, $\bb_\bsigma$.
Parameterizing the mean and variance of a Gaussian distribution based on a neural network gives us a natural and very expressive model for our latent space. It additionally comes with the benefit that we can use the \emph{reparameterization trick} \citep{Kingma_ICLR2014,Rezende_ICML2014} to backpropagate gradients of a loss function based on samples through the latent distribution.

\textbf{Generative model for $P_\btheta$.}
Using the approximate posterior distribution $Q_\bphi$ we generate observed samples (images) $\tilde{\bx}_t$ and $\tilde{\bx}_{t+1}$ from latent samples $\bz_t$ and $\bz_{t+1}$ by enforcing a locally linear relationship in latent space according to Equation \eqref{eq:dynamics_latent}, yielding the following generative model
\begin{equation}
\begin{array}{rll}
{\bz}_t &\sim \ Q_\bphi(Z \mid X) &= \ \mN(\bmu_t, \bSigma_t), \\
\hbz_{t+1} &\sim \ \hat{Q}_\bpsi(\hat{Z} \mid Z, \bu) \ &= \ \mN(\bA_t \bmu_t + \bB_t \bu_t + \bo_t, \bC_t), \\
\tbx_t, \tbx_{t+1} &\sim \ P_\btheta(X \mid Z) &= \ Bernoulli(\bp_t),
\end{array}
\label{eq:gen_model}
\end{equation}
where $\hat{Q}_\bpsi$ is the \emph{next latent state} posterior distribution, which exactly follows the linear form required for stochastic optimal control. With $\b\omega_t \sim \mathcal{N}(\mathbf{0}, \bH_t)$ as an estimate of the system noise, $\bC$ can be decomposed as $\bC_t = \bA_t \bSigma_t \bA^T_t + \bH_t $. 
Note that while the transition dynamics in our generative model operates on the inferred latent space, it takes untransformed controls into account. That is, we aim to learn a latent space such that the transition dynamics in $\bz$ linearizes the non-linear observed dynamics in $\bx$ and is locally linear in the applied controls $\bu$.
Reconstruction of an image from $\bz_t$ is performed by passing the sample through multiple hidden layers of a decoding neural network which computes the mean $\bp_t$ of the generative Bernoulli distribution\footnote{A Bernoulli distribution for $P_\btheta$ is a common choice when modeling black-and-white images.} $P_\btheta(X |Z)$ as
\begin{equation}
  {\bp_t} = \bW_\bp \dec(\bz_t) + \bb_\bp,
\end{equation}
where $\dec(\bz_t) \in \mathbb{R}^{n_d}$ is the response of the last hidden layer in the decoding network. The set of parameters for the decoding network, including weight matrix $\bW_\bp$ and bias $\bb_\bp$, then make up the learned generative parameters $\btheta$.

\textbf{Transition model for $\hat{Q}_\bpsi$.}
What remains is to specify how the linearization matrices $\bA_t \in \mathbb{R}^{n_z \times n_z}$, $\bB_t \in \mathbb{R}^{n_z \times n_u}$ and offset $\bo_t \in \mathbb{R}^{n_z}$ are predicted. Following the same approach as for distribution means and covariance matrices, we predict all local transformation parameters from samples $\bz_t$ based on the hidden representation $\trans(\bz_t) \in \mathbb{R}^{n_t}$ of a third neural network with parameters $\bpsi$ -- to which we refer as the transformation network. Specifically, we parametrize the transformation matrices and offset as
\begin{equation}
\begin{array}{rl}
\text{vec}[\bA_t] &= \bW_A \ \trans(\bz_t) + \bb_A, \\
\text{vec}[\bB_t] &= \bW_B \ \trans(\bz_t) + \bb_B, \\ 
\bo_t &= \bW_o \ \trans(\bz_t) + \bb_o,
\end{array}
\label{eq:trans_model}
\end{equation}
where $\text{vec}$ denotes vectorization and therefore $\text{vec}[\bA_t] \in \mathbb{R}^{(n_z^2)}$ and $\text{vec}[\bB_t] \in \mathbb{R}^{(n_z \cdot n_u)}$. To circumvent estimating the full matrix $\bA_t$ of size $n_z \times n_z$, we can choose it to be a perturbation of the identity matrix $\bA_t = (\bI + \bv_t \br_t^T)$ which reduces the parameters to be estimated for $\bA_t$ to $2n_z$.

A sketch of the complete architecture is shown in Figure \ref{fig:model}. It also visualizes an additional constraint that is essential for learning a representation for long-term predictions: we require samples $\hat{\bz}_{t+1}$ from the state transition distribution $\hat{Q}_\bpsi$ to be similar to the encoding of $\bx_{t+1}$ through $Q_\bphi$. 
While it might seem that just learning a perfect reconstruction of $\bx_{t+1}$ from $\hat{\bz}_{t+1}$ is enough, we require multi-step predictions for planning in $Z$ which \emph{must correspond} to valid trajectories in the observed space $X$. Without enforcing similarity between samples from $\hat{Q}_\bpsi$ and $Q_\bphi$, following a transition in latent space from $\bz_t$ with action $\bu_t$ may lead to a point $\hat{\bz}_{t+1}$, from which reconstruction of $\bx_{t+1}$ is possible, but that is not a valid encoding (i.e. the model will never encode any image as $\hat{\bz}_{t+1}$). Executing another action in $\hat{\bz}_{t+1}$ then does not result in a valid latent state -- since the transition model is conditional on samples coming from the inference network -- and thus long-term predictions fail. In a nutshell, such a divergence between encodings and the transition model results in a generative model that does not accurately model the Markov chain formed by the observations.

\subsection{Learning via stochastic gradient variational Bayes}
\label{sect:learning}
For training the model we use a data set $\mathcal{D} = \lbrace (\bx_1, \bu_1, \bx_{2}), \dots,  (\bx_{T-1}, \bu_{T-1}, \bx_{T}) \rbrace$ containing observation tuples with corresponding controls obtained from interactions with the dynamical system. 
Using this data set, we learn the parameters of the inference, transition and generative model by minimizing a variational bound on the true data negative log-likelihood $- \log P(\bx_t, \bu_t, \bx_{t+1})$ plus an additional constraint on the latent representation.
The complete loss function\footnote{Note that this is the loss for the latent state space model and distinct from the SOC costs.} is given as
\begin{equation}
  \cL(\mathcal{D}) = \sum_{(\bx_t, \bu_t, \bx_{t+1}) \in \mathcal{D}} \cL^{\text{bound}}(\bx_t, \bu_t, \bx_{t+1}) + \lambda \ \text{KL}\left (\hat{Q}_\bpsi(\hat{Z} \mid \bmu_t, \bu_t) \big \| Q_\bphi(Z \mid \bx_{t+1}) \right ).
\label{eq:loss}
\end{equation}
The first part of this loss is the per-example variational bound on the log-likelihood
\begin{equation}
\cL^{\text{bound}}(\bx_t, \bu_t, \bx_{t+1}) =  \mathbb{E}_{\substack{\bz_t \sim Q_\phi \\ \hat{\bz}_{t+1} \sim \hat{Q}_\psi}} \left [ -\log P_\btheta(\bx_t | \bz_t) -\log P_\btheta(\bx_{t+1} | \hat{\bz}_{t+1}) \right ] + \text{KL}(Q_\bphi || P(Z)),
\end{equation}
where $Q_{\bphi}$, $P_\btheta$ and $\hat{Q}_\bpsi$ are the parametric inference, generative and transition distributions from Section \ref{sect:model} and $P(Z_t)$ is a prior on the approximate posterior $Q_\bphi$; which we always chose to be an isotropic Gaussian distribution with mean zero and unit variance.
The second $\text{KL}$ divergence in Equation \eqref{eq:loss} is an additional contraction term with weight $\lambda$, that enforces agreement between the transition and inference models. This term is essential for establishing a Markov chain in latent space that corresponds to the real system dynamics (see Section \ref{sect:model} above for an in depth discussion). This KL divergence can also be seen as a prior on the latent transition model. Note that all KL terms can be computed analytically for our model (see supplementary for details). 

During training we approximate the expectation in $\cL(\mathcal{D})$ via sampling. Specifically, we take one sample $\bz_t$ for each input $\bx_t$ and transform that sample using Equation \eqref{eq:trans_model} to give a valid sample $\hat{\bz}_{t+1}$ from $\hat{Q}_\bpsi$.
We then jointly learn all parameters of our model by minimizing $\cL(\mathcal{D})$ using SGD. 

\section{Experimental Results}
We evaluate our model on four visual tasks: an agent in a plane with obstacles, a visual version of the classic inverted pendulum swing-up task, balancing a cart-pole system, and control of a three-link arm with larger images. These are described in detail below.

\subsection{Experimental Setup}
\textbf{Model training.}
We consider two different network types for our model: Standard fully connected neural networks with up to three layers, which work well for moderately sized images, are used for the planar and swing-up experiments; A deep convolutional network for the encoder in combination with an up-convolutional network as the decoder which, in accordance with recent findings from the literature \cite{Zeiler_CVPR2010,DosSprChair15}, we found to be an adequate model for larger images. Training was performed using Adam \cite{Kingma_2015} throughout all experiments. 
The training data set $\mathcal{D}$ for all tasks was generated by randomly sampling $N$ state observations and actions with corresponding successor states. For the plane we used $N \!\! = \!\! 3,000$ samples, for the inverted pendulum and cart-pole system we used $N \!\! = \!\! 15,000$ and for the arm $N \!\! = \!\! 30,000$.
A complete list of architecture parameters and hyperparameter choices as well as an in-depth explanation of the up-convolutional network are specified in the supplementary material.
We will make our code and a video containing controlled trajectories for all systems available under \url{http://ml.informatik.uni-freiburg.de/research/e2c} .

\textbf{Model variants.}
In addition to the Embed to Control (E2C) dynamics model derived above, we also consider two variants:
By removing the latent dynamics network $\trans$, i.e. setting its output to one in Equation \eqref{eq:trans_model} -- we obtain a variant in which $\bA_t$, $\bB_t$ and $\bo_t$ are estimated as globally linear matrices (Global E2C). If we instead replace the transition model with a network estimating the dynamics as a non-linear function $\hat{f}^{\text{lat}}$  and only linearize during planning, estimating $\bA_t$, $\bB_t$, $\bo_t$ as Jacobians to $\hat{f}^{\text{lat}}$ as described in Section \ref{sect:soc}, we obtain a variant with nonlinear latent dynamics. 

\textbf{Baseline models.}
For a thorough comparison and to exhibit the complicated nature of the 
tasks, we also test a set of baseline models on the plane and the inverted pendulum task (using the same architecture as the E2C model): a standard variational autoencoder (VAE) and a deep autoencoder (AE) are trained on the autoencoding subtask for visual problems. That is, given a data set $\mathcal{D}$ used for training our model, we remove all actions from the tuples in $\mathcal{D}$ and disregard temporal context between images. After autoencoder training we learn a dynamics model in latent space, approximating $f^{\text{lat}}$ from Section \ref{sect:soc}.
We also consider a VAE variant with a slowness term on the latent representation -- a full description of this variant is given in the supplementary material.

\textbf{Optimal control algorithms.}
To perform optimal control in the latent space of different models, we employ two trajectory optimization algorithms: iterative linear quadratic regulation (iLQR) \cite{Li_ICINC2004} (for the plane and inverted pendulum) and approximate inference control (AICO) \cite{Toussaint_ICML2009} (all other experiments). For all VAEs both methods operate on the mean of distributions $Q_\bphi$ and $\hat{Q}_\bpsi$. AICO additionally makes use of the local Gaussian covariances $\bSigma_t$ and $\bC_t$. Except for the experiments on the planar system, control was performed in a model predictive control fashion using the receding horizon scheme introduced in \cite{Tassa_2008}. To obtain closed loop control given an image $\bx_t$, it is first passed through the encoder to obtain the latent state $\bz_t$. A locally optimal trajectory is subsequently found by optimizing $(\bz^*_{t:t+T}, \bu^*_{t:t+T}) \approx \arg \min_{\substack{\bz_{t:t+T} \\ \bu_{t:t+T}}} J(\bz_{t:t+T}, \bu_{t:t+T})$ with fixed, small horizon $T$ (with $T = 10$ unless noted otherwise). Controls $\bu^*_t$ are applied to the system and a transition to $\bz_{t+1}$ is observed (by encoding the next image $\bx_{t+1}$). Then a new control sequence -- with horizon $T$ -- starting in $\bz_{t+1}$ is found using the last estimated trajectory as a bootstrap. Note that planning is performed entirely in the latent state \emph{without access to any observations} except for the depiction of the current state. To compute the cost function $c(\bz_t, \bu_t)$ required for trajectory optimization in $\bz$ we assume knowledge of the observation $\bx_{\text{goal}}$ of the goal state $\bs_{\text{goal}}$. This observation is then transformed into latent space and costs are computed according to Equation \eqref{eq:costs}.

\subsection{Control in a planar system}
The agent in the planar system can move in a bounded two-dimensional plane by choosing a continuous offset in x- and y-direction. The high-dimensional representation of a state is a $40 \times 40$ black-and-white image.
Obstructed by six circular obstacles, the task is to move to the bottom right of the image, starting from a random x position at the top of the image. 
The encodings of obstacles are obtained prior to planning and an additional quadratic cost term is penalizing proximity to them.
 
\begin{figure}[t]
	\centering
	\input{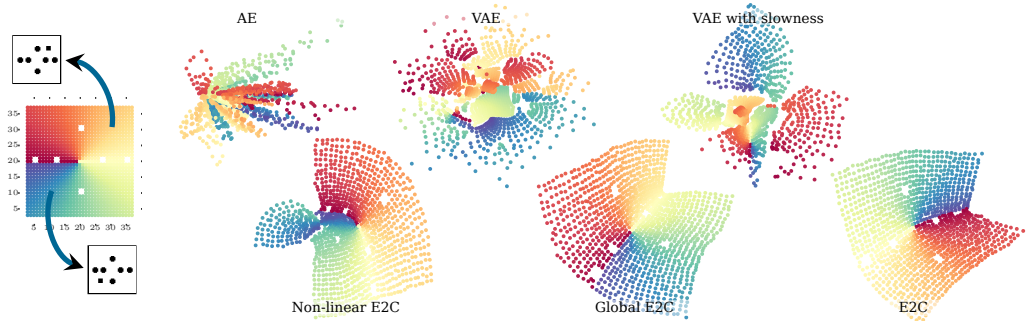}
        \vspace{-0.4cm}
	\caption{The true state space of the planar system (left) with examples (obstacles encoded as circles) and the inferred spaces (right) of different models. The spaces are spanned by generating images for every valid position of the agent and embedding them with the respective encoders.}
        \vspace{-0.5cm}
	\label{fig:plane_latent}
\end{figure}

A depiction of the observations on which control is performed -- together with their corresponding state values and embeddings into latent space -- is shown in Figure \ref{fig:plane_latent}. The figure also clearly shows a fundamental advantage the E2C model has over its competitors: While the separately trained autoencoders make for aesthetically pleasing pictures, the models failed to discover the underlying structure of the state space, complicating dynamics estimation and largely invalidating costs based on distances in said space. Including the latent dynamics constraints in these end-to-end models on the other hand, yields latent spaces approaching the optimal planar embedding. 

We test the long-term accuracy by accumulating latent and real trajectory costs to quantify whether the imagined trajectory reflects reality.
The results for all models when starting from random positions at the top and executing $40$ pre-computed actions are summarized in Table \ref{tab:results} -- using a seperate test set for evaluating reconstructions. While all methods achieve a low reconstruction loss, the difference in accumulated real costs per trajectory show the superiority of the E2C model. Using the globally or locally linear E2C model, trajectories planned in latent space are as good as trajectories planned on the real state. All models besides E2C fail to give long-term predictions that result in good performance.

\begin{table}[t]
\begin{center}
\caption{Comparison between different approaches to model learning
  from raw pixels for the planar and pendulum system. We compare all models with
  respect to their prediction quality on a test set of sampled
  transitions and with respect to their performance when combined with SOC (trajectory cost for control from different start states). Note that trajectory costs in latent space are not necessarily comparable. The ``real'' trajectory cost was computed on the dynamics of the simulator while executing planned actions. For the true models for $\bs_t$, real trajectory costs were $20.24 \pm 4.15$ for the planar system, and $9.8 \pm 2.4$ for the pendulum. Success was defined as reaching the goal state and staying $\epsilon$-close to it for the rest of the trajectory (if non terminating). All statistics quantify over 5/30 (plane/pendulum) different starting positions. A $\dagger$ marks separately trained dynamics networks.}
\vspace{0.15cm}
\footnotesize
\begin{tabular}{lccccc}
\multicolumn{1}{c}{\bf Algorithm}  &\multicolumn{1}{c}{\bf State Loss} & \bf Next State Loss & \multicolumn{2}{c}{\bf Trajectory Cost} & \bf Success \\
                                   & \bf log $\mathbf{p(\bx_t | \hat{\bx}_t)}$ & \bf log $\mathbf{p(x_{t+1}|\hat{x}_t, u_t)}$ & \bf Latent & \bf Real & \textbf{percent} \\ 
\hline
\multicolumn{6}{c}{\bf Planar System} \\
AE$^\dagger$ & 11.5 $\pm$ 97.8 & 3538.9 $\pm$ 1395.2 & 1325.6 $\pm$ 81.2  & 273.3  $\pm$ 16.4 & 0 $\%$ \\
VAE$^\dagger$ & 3.6 $\pm$ 18.9 & 652.1 $\pm$ 930.6 & 43.1 $\pm$ 20.8 & 91.3 $\pm$ 16.4 & 0 $\%$  \\
VAE + slowness$^\dagger$ & 10.5 $\pm$ 22.8 & 104.3 $\pm$ 235.8 & 47.1 $\pm$ 20.5 & 89.1 $\pm$ 16.4 & 0 $\%$  \\
Non-linear E2C & 8.3 $\pm$ 5.5 & 11.3 $\pm$ 10.1 & 19.8 $\pm$ 9.8 & 42.3 $\pm$ 16.4 &  96.6 $\%$ \\
Global E2C & \textbf{6.9 $\pm$ 3.2} & \textbf{9.3 $\pm$ 4.6} & 12.5 $\pm$ 3.9 & 27.3 $\pm$ 9.7 & \textbf{100 $\%$} \\
\bfseries E2C & 7.7 $\pm$ 2.0 & 9.7 $\pm$ 3.2 & 10.3 $\pm$ 2.8 & \textbf{25.1 $\pm$ 5.3} & \textbf{100 $\%$} \\ \hline

\multicolumn{6}{c}{\bf Inverted Pendulum Swing-Up} \\
AE$^\dagger$ & 8.9 $\pm$ 100.3 & 13433.8 $\pm$ 6238.8 & 1285.9 $\pm$ 355.8 & 194.7 $\pm$ 44.8 &  0 $\%$ \\
VAE$^\dagger$  & 7.5 $\pm$ 47.7 & 8791.2 $\pm$ 17356.9 & 497.8 $\pm$ 129.4 & 237.2 $\pm$ 41.2 &  0 $\%$ \\
VAE + slowness$^\dagger$ & 26.5 $\pm$ 18.0 & 779.7 $\pm$ 633.3 & 419.5 $\pm$ 85.8 & 188.2 $\pm$ 43.6 & 0 $\%$ \\
E2C no latent KL & 64.4 $\pm$ 32.8  & 87.7 $\pm$ 64.2  & 489.1 $\pm$ 87.5 & 213.2 $\pm$ 84.3 &  0 $\%$ \\
Non-linear E2C & \textbf{59.6 $\pm$ 25.2} & \textbf{72.6 $\pm$ 34.5} & 313.3 $\pm$ 65.7 & 37.4 $\pm$ 12.4 & 63.33 $\%$ \\
Global E2C & 115.5 $\pm$ 56.9 & 125.3 $\pm$ 62.6 & 628.1 $\pm$ 45.9 & 125.1 $\pm$ 10.7 & 0 $\%$ \\
\textbf{E2C} & 84.0 $\pm$ 50.8 & 89.3 $\pm$ 42.9 & 275.0 $\pm$ 16.6 & \textbf{15.4 $\pm$ 3.4} & \textbf{90 $\%$} \\
\end{tabular}
\label{tab:results}
\end{center}
\vspace{-0.5cm}
\end{table}
 
\subsection{Learning swing-up for an inverted pendulum}
\begin{figure}[b]
 	\centering
    \input{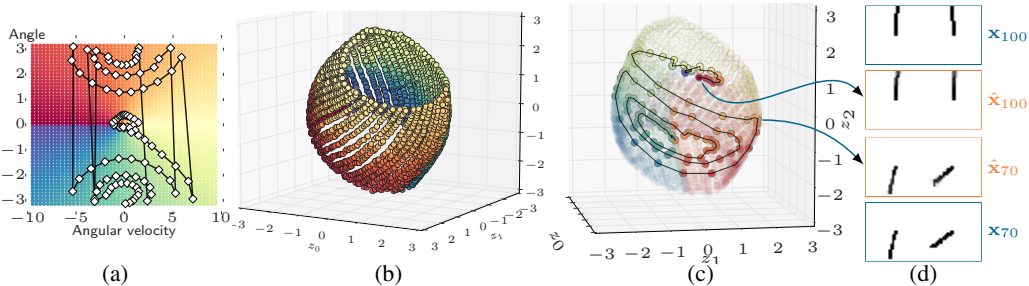}
    \vspace{-0.3cm}
    \caption{(a) The true state space of the inverted pendulum task overlaid with a successful trajectory taken by the E2C agent.
             (b) The learned latent space.
             (c) The trajectory from (a) traced out in the latent space.
             (d) Images $\bx$ and reconstructions $\hat{\bx}$ showing current positions (right) and history (left).}
           \label{fig:pole_latent}
    \vspace{-0.2cm}
\end{figure}

We next turn to the task of controlling the classical inverted pendulum system \cite{Wang_TFS1996} from images. We create depictions of the state by rendering a fixed length line starting from the center of the image at an angle corresponding to the pendulum position. The goal in this task is to swing-up and balance an underactuated pendulum from a resting position (pendulum hanging down). Exemplary observations and reconstructions for this system are given in Figure \ref{fig:pole_latent}(d).
In the visual inverted pendulum task our algorithm faces two additional difficulties: the observed space is non-Markov, as the angular velocity cannot be inferred from a single image, and second, discretization errors due to rendering pendulum angles as small 48x48 pixel images make exact control difficult. To restore the Markov property, we stack two images (as input channels), thus observing a one-step history. 

Figure \ref{fig:pole_latent} shows the topology of the latent space for our model, as well as one sample trajectory in true state and latent space. 
The fact that the model can learn a meaningful embedding, separating velocities and positions, from this data is remarkable (no other model recovered this shape). Table \ref{tab:results} again compares the different models quantitatively. While the E2C model is not the best in terms of reconstruction performance, it is the only model resulting in stable swing-up \textit{and} balance behavior. 
We explain the failure of the other models with the fact that the non-linear latent dynamics model cannot be guaranteed to be linearizable for all control magnitudes, resulting in undesired behavior around unstable fixpoints of the real system dynamics, and that for this task a globally linear dynamics model is inadequate.

\subsection{Balancing a cart-pole and controlling a simulated robot arm}
Finally, we consider control of two more complex dynamical systems from images using a six layer convolutional inference and six layer up-convolutional generative network, resulting in a \emph{12-layer deep} path from input to reconstruction. Specifically, we control a visual version of the classical cart-pole system \cite{Sutton_Barto_1998} from a history of two $80 \times 80$ pixel images as well as a three-link planar robot arm based on a history of two $128 \times 128$ pixel images. The latent space was set to be 8-dimensional in both experiments. The real state dimensionality for the cart-pole is four and is controlled using one action, while for the arm the real state can be described in 6 dimensions (joint angles and velocities) and controlled using a three-dimensional action vector corresponding to motor torques. 

\begin{figure}[t]
 	\centering
        \includegraphics[width=0.8\textwidth]{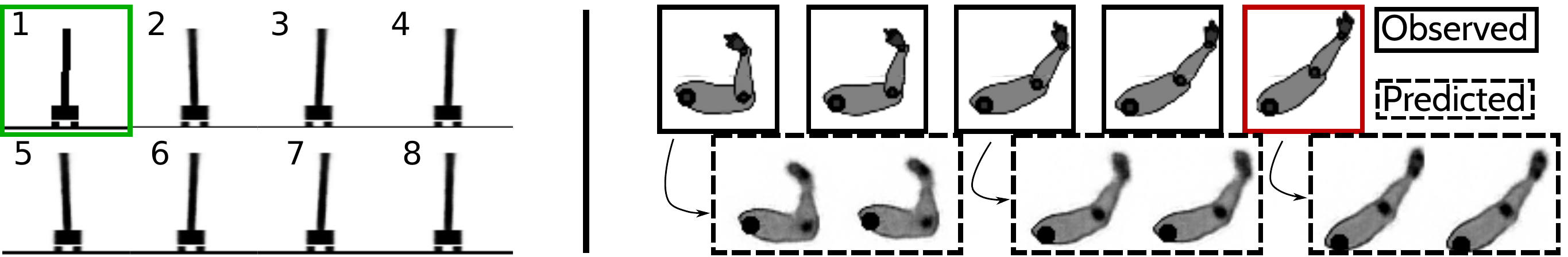}
        \vspace{-0.3cm}
        \caption{Left: Trajectory from the cart-pole domain. Only the first image (green) is ``real'', all other images are ``dreamed up'' by our model. Notice discretization artifacts present in the real image. Right: Exemplary observed (with history image omitted) and predicted images (including the history image) for a trajectory in the visual robot arm domain with the goal marked in red.}
        \label{fig:cart_and_robot}
        \vspace*{-0.45cm}
\end{figure}

As in previous experiments the E2C model seems to have no problem finding a locally linear embedding of images into latent space in which control can be performed. Figure \ref{fig:cart_and_robot} depicts exemplary images -- for both problems -- from a trajectory executed by our system. The costs for these trajectories ($11.13$ for the cart-pole, $85.12$ for the arm) are only slightly worse than trajectories obtained by AICO operating on the real system dynamics starting from the same start-state ($7.28$ and $60.74$ respectively). The supplementary material contains additional experiments using these domains.

\section{Comparison to recent work}
In the context of representation learning for control (see \citet{Boehmer_KI2015} for a review), deep autoencoders (ignoring state transitions) similar to our baseline models have been applied previously, e.g. by \citet{Lange10}.
A more direct route to control based on image streams is taken by recent work on (model free) deep end-to-end Q-learning for Atari games by \citet{Mnih_Nature2015}, as well as kernel based \cite{Hoof_2015} and deep policy learning for robot control \cite{Levine_arxiv2015}.

Close to our approach is a recent paper by \citet{Wahlstrom_arxiv2015}, where autoencoders are used to extract a latent representation for control from images, on which a non-linear model of the forward dynamics is learned. Their model is trained jointly and is thus similar to the non-linear E2C variant in our comparison. In contrast to our model, their formulation requires PCA pre-processing and does neither ensure that long-term predictions in latent space do not diverge, nor that they are linearizable. 

As stated above, our system belongs to the family of VAEs and is generally similar to recent
work such as \citet{Kingma_ICLR2014,Rezende_ICML2014,Gregor_ICML2015,Bayer2014b}.
Two additional parallels between our work and recent advances for training deep neural networks can be observed. First, the idea of enforcing desired transformations in latent space during learning -- such that the data becomes easy to model -- has appeared several times already in the literature. This includes the development of transforming auto-encoders \cite{Hinton2011} and recent probabilistic models for images \cite{Dinh_2015,Cohen_ICLR2015}. Second, learning relations between pairs of images -- although \emph{without control} -- has received considerable attention from the community during the last years \cite{Taylor2010,Memisevic2013}.
In a broader context our model is related to work on state estimation in Markov decision processes (see \citet{LangfordSalZhang09} for a discussion) through, e.g., hidden Markov models and Kalman filters \cite{WestHarr97,Matsubara2014}. 

\section{Conclusion}

We presented Embed to Control (E2C), a system for stochastic optimal control on high-dimensional image streams. Key to the approach is the extraction of a latent dynamics model which is constrained to be locally linear in its state transitions. An evaluation on four challenging benchmarks revealed that E2C can find embeddings on which control can be performed with ease, reaching performance close to that achievable by optimal control on the real system model.

\subsubsection*{Acknowledgments}
We thank A. Radford, L. Metz, and T. DeWolf for sharing code, as well as A. Dosovitskiy for useful discussions. This work was partly funded by a DFG grant within the priority program ``Autonomous learning'' (SPP1597) and the BrainLinks-BrainTools Cluster of Excellence (grant number EXC 1086). M. Watter is funded through the State Graduate Funding Program of Baden-W\"urttemberg.

\bibliographystyle{my-unsrtnat}
{
\small
\bibliography{linearizing}

\begin{thebibliography}{39}
\providecommand{\natexlab}[1]{#1}
\providecommand{\url}[1]{\texttt{#1}}
\expandafter\ifx\csname urlstyle\endcsname\relax
  \providecommand{\doi}[1]{doi: #1}\else
  \providecommand{\doi}{doi: \begingroup \urlstyle{rm}\Url}\fi

\bibitem[Jacobson and Mayne(1970)]{Jacobson1970}
D.~Jacobson and D.~Mayne.
\newblock Differential dynamic programming.
\newblock \emph{American Elsevier}, 1970.

\bibitem[Todorov and Li(2005)]{Todorov2005}
E.~Todorov and W.~Li.
\newblock A generalized iterative {LQG} method for locally-optimal feedback
  control of constrained nonlinear stochastic systems.
\newblock In \emph{{ACC}}. IEEE, 2005.

\bibitem[Tassa et~al.(2008)Tassa, Erez, and Smart]{Tassa_2008}
Y.~Tassa, T.~Erez, and W.~D. Smart.
\newblock Receding horizon differential dynamic programming.
\newblock In \emph{{Proc. of NIPS}}, 2008.

\bibitem[Pan and Theodorou(2014)]{Pan_2014}
Y.~Pan and E.~Theodorou.
\newblock Probabilistic differential dynamic programming.
\newblock In \emph{{Proc. of NIPS}}, 2014.

\bibitem[Levine and Koltun(2013)]{Levine_2013n}
S.~Levine and V.~Koltun.
\newblock Variational policy search via trajectory optimization.
\newblock In \emph{{Proc. of NIPS}}, 2013.

\bibitem[Kingma and Welling(2014)]{Kingma_ICLR2014}
D.~P. Kingma and M.~Welling.
\newblock Auto-encoding variational bayes.
\newblock In \emph{{Proc. of ICLR}}, 2014.

\bibitem[Rezende et~al.(2014)Rezende, Mohamed, and Wierstra]{Rezende_ICML2014}
D.~J. Rezende, S.~Mohamed, and D.~Wierstra.
\newblock Stochastic backpropagation and approximate inference in deep
  generative models.
\newblock In \emph{{Proc. of ICML}}, 2014.

\bibitem[Zeiler et~al.(2010)Zeiler, Krishnan, Taylor, and
  Fergus]{Zeiler_CVPR2010}
M.~D. Zeiler, D.~Krishnan, G.~W. Taylor, and R.~Fergus.
\newblock Deconvolutional networks.
\newblock In \emph{{CVPR}}, 2010.

\bibitem[Dosovitskiy et~al.(2015)Dosovitskiy, Springenberg, and
  Brox]{DosSprChair15}
A.~Dosovitskiy, J.~T. Springenberg, and T.~Brox.
\newblock Learning to generate chairs with convolutional neural networks.
\newblock In \emph{{{Proc. of CVPR}}}, 2015.

\bibitem[Stengel(1994)]{Stengel_Book1994}
R.~F. Stengel.
\newblock \emph{Optimal Control and Estimation}.
\newblock Dover Publications, 1994.

\bibitem[Li and Todorov(2004)]{Li_ICINC2004}
W.~Li and E.~Todorov.
\newblock {Iterative Linear Quadratic Regulator Design for Nonlinear Biological
  Movement Systems}.
\newblock In \emph{{Proc. of ICINCO}}, 2004.

\bibitem[Toussaint(2009)]{Toussaint_ICML2009}
M.~Toussaint.
\newblock {Robot Trajectory Optimization using Approximate Inference}.
\newblock In \emph{{Proc. of ICML}}, 2009.

\bibitem[Jordan et~al.(1999)Jordan, Ghahramani, Jaakkola, and Saul]{Jordan1999}
M.~I. Jordan, Z.~Ghahramani, T.~S. Jaakkola, and L.~K. Saul.
\newblock An introduction to variational methods for graphical models.
\newblock In \emph{Machine Learning}, 1999.

\bibitem[Kingma and Ba(2015)]{Kingma_2015}
D.~Kingma and J.~Ba.
\newblock Adam: A method for stochastic optimization.
\newblock In \emph{{Proc. of ICLR}}, 2015.

\bibitem[Wang et~al.(1996)Wang, Tanaka, and Griffin]{Wang_TFS1996}
H.~Wang, K.~Tanaka, and M.~Griffin.
\newblock An approach to fuzzy control of nonlinear systems; stability and
  design issues.
\newblock \emph{IEEE Trans. on Fuzzy Systems}, 4\penalty0 (1), 1996.

\bibitem[Sutton and Barto(1998)]{Sutton_Barto_1998}
R.~S. Sutton and A.~G. Barto.
\newblock \emph{Introduction to Reinforcement Learning}.
\newblock MIT Press, Cambridge, MA, USA, 1st edition, 1998.
\newblock ISBN 0262193981.

\bibitem[B{\"{o}}hmer et~al.(2015)B{\"{o}}hmer, Springenberg, Boedecker,
  Riedmiller, and Obermayer]{Boehmer_KI2015}
W.~B{\"{o}}hmer, J.~T. Springenberg, J.~Boedecker, M.~Riedmiller, and
  K.~Obermayer.
\newblock Autonomous learning of state representations for control.
\newblock \emph{KI - K{\"{u}}nstliche Intelligenz}, 2015.

\bibitem[Lange and Riedmiller(2010)]{Lange10}
S.~Lange and M.~Riedmiller.
\newblock Deep auto-encoder neural networks in reinforcement learning.
\newblock In \emph{{Proc. of IJCNN}}, 2010.

\bibitem[Mnih et~al.(2015)Mnih, Kavukcuoglu, Silver, Rusu, Veness, Bellemare,
  Graves, Riedmiller, Fidjeland, Ostrovski, Petersen, Beattie, Sadik,
  Antonoglou, King, Kumaran, Wierstra, Legg, and Hassabis]{Mnih_Nature2015}
V.~Mnih, K.~Kavukcuoglu, D.~Silver, A.~A. Rusu, J.~Veness, M.~G. Bellemare,
  A.~Graves, M.~Riedmiller, A.~K. Fidjeland, G.~Ostrovski, S.~Petersen,
  C.~Beattie, A.~Sadik, I.~Antonoglou, H.~King, D.~Kumaran, D.~Wierstra,
  S.~Legg, and D.~Hassabis.
\newblock Human-level control through deep reinforcement learning.
\newblock \emph{Nature}, 518\penalty0 (7540), 02 2015.

\bibitem[van Hoof et~al.(2015)van Hoof, Peters, and Neumann]{Hoof_2015}
H.~van Hoof, J.~Peters, and G.~Neumann.
\newblock Learning of non-parametric control policies with high-dimensional
  state features.
\newblock In \emph{Proc. of AISTATS}, 2015.

\bibitem[Levine et~al.(2015)Levine, Finn, Darrell, and
  Abbeel]{Levine_arxiv2015}
S.~Levine, C.~Finn, T.~Darrell, and P.~Abbeel.
\newblock End-to-end training of deep visuomotor policies.
\newblock \emph{CoRR}, abs/1504.00702, 2015.
\newblock URL \url{http://arxiv.org/abs/1504.00702}.

\bibitem[Wahlstr{\"{o}}m et~al.(2015)Wahlstr{\"{o}}m, Sch{\"{o}}n, and
  Deisenroth]{Wahlstrom_arxiv2015}
N.~Wahlstr{\"{o}}m, T.~B. Sch{\"{o}}n, and M.~P. Deisenroth.
\newblock From pixels to torques: Policy learning with deep dynamical models.
\newblock \emph{CoRR}, abs/1502.02251, 2015.
\newblock URL \url{http://arxiv.org/abs/1502.02251}.

\bibitem[Gregor et~al.(2015)Gregor, Danihelka, Graves, Rezende, and
  Wierstra]{Gregor_ICML2015}
K.~Gregor, I.~Danihelka, A.~Graves, D.~Rezende, and D.~Wierstra.
\newblock {DRAW}: A recurrent neural network for image generation.
\newblock In \emph{{Proc. of ICML}}, 2015.

\bibitem[Bayer and Osendorfer(2014)]{Bayer2014b}
J.~Bayer and C.~Osendorfer.
\newblock Learning stochastic recurrent networks.
\newblock In \emph{NIPS 2014 Workshop on Advances in Variational Inference},
  2014.

\bibitem[Hinton et~al.(2011)Hinton, Krizhevsky, and Wang]{Hinton2011}
G.~Hinton, A.~Krizhevsky, and S.~Wang.
\newblock Transforming auto-encoders.
\newblock In \emph{Proc. of ICANN}, 2011.

\bibitem[Dinh et~al.(2015)Dinh, Krueger, and Bengio]{Dinh_2015}
L.~Dinh, D.~Krueger, and Y.~Bengio.
\newblock Nice: Non-linear independent components estimation.
\newblock \emph{CoRR}, abs/1410.8516, 2015.
\newblock URL \url{http://arxiv.org/abs/1410.8516}.

\bibitem[Cohen and Welling(2015)]{Cohen_ICLR2015}
T.~Cohen and M.~Welling.
\newblock Transformation properties of learned visual representations.
\newblock In \emph{{ICLR}}, 2015.

\bibitem[Taylor et~al.(2010)Taylor, Sigal, Fleet, and Hinton]{Taylor2010}
G.~W. Taylor, L.~Sigal, D.~J. Fleet, and G.~E. Hinton.
\newblock Dynamical binary latent variable models for 3d human pose tracking.
\newblock In \emph{Proc. of CVPR}, 2010.

\bibitem[Memisevic(2013)]{Memisevic2013}
R.~Memisevic.
\newblock Learning to relate images.
\newblock \emph{IEEE Trans. on PAMI}, 35\penalty0 (8):\penalty0 1829--1846,
  2013.

\bibitem[Langford et~al.(2009)Langford, Salakhutdinov, and
  Zhang]{LangfordSalZhang09}
J.~Langford, R.~Salakhutdinov, and T.~Zhang.
\newblock Learning nonlinear dynamic models.
\newblock In \emph{ICML}, 2009.

\bibitem[West and Harrison(1997)]{WestHarr97}
M.~West and J.~Harrison.
\newblock \emph{Bayesian Forecasting and Dynamic Models (Springer Series in
  Statistics)}.
\newblock Springer-Verlag, February 1997.
\newblock ISBN 0387947256.

\bibitem[Matsubara et~al.(2014)Matsubara, G\'omez, and Kappen]{Matsubara2014}
T.~Matsubara, V.~G\'omez, and H.~J. Kappen.
\newblock Latent {Kullback Leibler} control for continuous-state systems using
  probabilistic graphical models.
\newblock \emph{UAI}, 2014.

\bibitem[Kulkarni et~al.(2015)Kulkarni, Whitney, Kohli, and
  Tenenbaum]{Kulkarni_arxiv2015}
T.~D. Kulkarni, W.~Whitney, P.~Kohli, and J.~B. Tenenbaum.
\newblock Deep convolutional inverse graphics network.
\newblock \emph{CoRR}, abs/1503.03167, 2015.
\newblock URL \url{http://arxiv.org/abs/1503.03167}.

\bibitem[Osendorfer et~al.(2014)Osendorfer, Soyer, and van~der
  Smagt]{OseSoySma2014}
C.~Osendorfer, H.~Soyer, and P.~van~der Smagt.
\newblock Image super-resolution with fast approximate convolutional sparse
  coding.
\newblock In \emph{{Proc. of ICONIP}}, Lecture Notes in Computer Science.
  Springer International Publishing, 2014.

\bibitem[Jonschkowski and Brock(2014)]{Jonschkowski14}
R.~Jonschkowski and O.~Brock.
\newblock State representation learning in robotics: Using prior knowledge
  about physical interaction.
\newblock In \emph{{Proc. of RSS}}, 2014.

\bibitem[Legenstein et~al.(2010)Legenstein, Wilbert, and Wiskott]{Legenstein10}
R.~Legenstein, N.~Wilbert, and L.~Wiskott.
\newblock Reinforcement learning on slow features of high-dimensional input
  streams.
\newblock \emph{PLoS Computational Biology}, 2010.

\bibitem[Zou et~al.(2011)Zou, Ng, and Yu]{NIPS2011Zou}
W.~Zou, A.~Ng, and K.~Yu.
\newblock Unsupervised learning of visual invariance with temporal coherence.
\newblock In \emph{NIPS*2011 Workshop on Deep Learning and Unsupervised Feature
  Learning}, 2011.

\bibitem[Saxe et~al.(2014)Saxe, McClelland, and Ganguli]{Saxe13}
A.~M. Saxe, J.~L. McClelland, and S.~Ganguli.
\newblock Exact solutions to the nonlinear dynamics of learning in deep linear
  neural networks.
\newblock In \emph{{Proc. of ICLR}}, 2014.

\bibitem[Glorot et~al.(2011)Glorot, Bordes, and Bengio]{GlorotBB11}
X.~Glorot, A.~Bordes, and Y.~Bengio.
\newblock Deep sparse rectifier neural networks.
\newblock In \emph{{AISTATS}}. Journal of Machine Learning Research - Workshop
  and Conference Proceedings, 2011.

\end{thebibliography}
}

\begin{appendix}
\section{Supplementary to the E2C description}
\label{sect:description}

\subsection{State transition matrix factorization and KL Divergence}

As alluded to in the main paper, estimation of the full local state transition matrix $\bA_t \in \mathbb{R}^{n_z \times n_z}$ from Equation (8) requires the transition network to predict $n_z \times n_z$ parameters. Using an arbitrary state transition matrix also -- inconveniently -- requires inversion of said matrix for computing the KL divergence penalty from Equation (11) (through which it is hard to backpropagate). We started our experiments using a full matrix (and only approximating all KL divergence terms), but quickly found that a rank one pertubation of the identity matrix could be used instead without loss of performance in any of our benchmarks. To the contrary, the resulting networks have fewer parameters and are thus easier to train.
We here give the derivation of this process and how the KL divergence from Equation (11) can be computed.
For the reformulation we represent $\bA_t$ as $\bA_t = \bI + \bv_t \br_t^T$, therefore only $\bv_t$ and $\br_t$ need to be estimated by the transition network, reducing the number of outputs for $\bA_t$ from $n^2_z$ to $2n_z$. 

The KL divergence between two multivariate Gaussians is given by
\begin{align}
\text{KL}(\mathcal{N}_0 || \mathcal{N}_1) = \frac{1}{2} \left( \Tr \left(\bSigma^{-1}_1 \bSigma_0 \right) + (\bmu_1 - \bmu_0)^T \bSigma^{-1}_1 (\bmu_1 - \bmu_0) - k + \log \left( \frac{\det \bSigma_1}{\det \bSigma_0} \right) \right).
\end{align}

For a simplified notation, such that $\text{KL}(\mathcal{N}_0 || \mathcal{N}_1) = \text{KL}(\hat{Q} || Q)$, let us assume 
\begin{align*}
    \mathcal{N}_0 & = \mathcal{N}(\bmu_0, \bA \bSigma_0 \bA^T) = \mathcal{N}(\bmu_{t}, \bA_t \bSigma_{t} \bA_t^T) = \hat{Q}, \\
    \mathcal{N}_1 & = \mathcal{N}(\bmu_1, \bSigma_1) = \mathcal{N}(\bmu_{t+1}, \bSigma_{t+1}) = Q.
\end{align*}

The main point behind the derivation presented in the following, is to make partial derivatives of the above KL divergence efficiently computable. To this end, we cannot take the trace or the determinant via numerical algorithms, because we have to be able to take the gradients in symbolic form. Aside from that, we like to process a batch of samples, so the computation should have a convenient form and not require excessive amounts of tensor products in between. We start our simplification with the trace term which results in
\begin{align*}
\Tr \left(\bSigma^{-1}_1 \bSigma_0 \right) & = \Tr \left(\bSigma^{-1}_1 \bA \bSigma_0 \bA^T \right) \\
& = \Tr \left(\bSigma^{-1}_1 (\mathbf{I} + \bv \br^T) \bSigma_0 (\mathbf{I} + \bv \br^T)^T \right) \\
& = \Tr \left( \left( \bSigma^{-1}_1 + \bSigma^{-1}_1 \bv \br^T \right) \left( \bSigma_0 + \bSigma_0 (\bv \br^T)^T \right) \right) \\
& = \Tr \left( \bSigma^{-1}_1 \bSigma_0 + \bSigma^{-1}_1 \bSigma_0 (\bv \br^T)^T + \bSigma^{-1}_1 \bv \br^T \bSigma_0 + \bSigma^{-1}_1 \bv \br^T \bSigma_0 (\bv \br^T)^T \right) \tag*{\tiny$\Tr(A + B) = \Tr(A) + \Tr(B) $} \\
& = \Tr \left( \bSigma^{-1}_1 \bSigma_0 \right) + \Tr \left( \bSigma^{-1}_1 \bSigma_0 (\bv \br^T)^T \right) + \Tr \left( \bSigma^{-1}_1 \bv \br^T \bSigma_0 \right) + \Tr \left( \bSigma^{-1}_1 \bv \br^T \bSigma_0 \br \bv^T \right) \tag*{\tiny$\Tr(ABC) = \Tr(CAB) = \ldots$} \\
& = \sum_i \frac{\sigma_{0, i}^2}{\sigma_{1, i}^2} + \sum_i \frac{\sigma_{0, i}^2 r_i v_i}{\sigma_{1, i}^2} + \sum_i \frac{v_i r_i \sigma_{0, i}^2}{\sigma_{1, i}^2} + \Tr \left(\bv^T \bSigma^{-1}_1 \bv \br^T \bSigma_0 \br \right)\\
& = \sum_i \frac{\sigma_{0, i}^2 + 2 \sigma_{0, i}^2 v_i r_i}{\sigma_{1, i}^2} + \sum_i r_i^2 \sigma_i^2 \cdot \sum_i \frac{v_i^2}{\sigma_i^2}. \\
\end{align*}
The last equation is easy to implement and only requires summing over the non-batch dimension. The difference of means can be derived very quickly with the same summing scheme:
\begin{align*}
(\bmu_1 - \bmu_0)^T \bSigma^{-1}_1 (\bmu_1 - \bmu_0) = \sum_i \frac{(\bmu_1 - \bmu_0)_i^2}{\sigma_{1, i}^2}.
\end{align*}

It remains the ratio of determinants, which we will simplify with the matrix determinant lemma giving
\begin{align*}
\log \left( \frac{\det \bSigma_1}{\det \bA \bSigma_0 \bA^T} \right) & = \log \det \bSigma_1 - \log \det \left( \bA \bSigma_0 \bA^T \right) \\
& = \log \prod_i \sigma_{1, i}^2 - \log \left(\det \bA \cdot \det \bSigma_0 \cdot \det \bA^T \right) \tag*{\tiny$\det \bA^T = \det \bA$} \\
& = 2 \sum_i \log \sigma_{1, i} - \log \left( (\det \bA)^2 \prod_i \sigma_{0, i}^2 \right) \tag*{\tiny $\text{Matrix determinant lemma}$}\\
& = 2 \sum_i \log \sigma_{1, i} - \log \left(1 + \bv^T \br \right)^2 - 2 \sum_i \log \sigma_{0, i} \\
& = 2 \left( \sum_i \left( \log \sigma_{1, i}^2 - \log \sigma_{0, i}^2 \right) - \log (1 + \sum_i v_i r_i) \right).
\end{align*}
Putting the above to formulas together finally yields
\begin{alignat}{3}
    \text{KL}(\mathcal{N}_0 || \mathcal{N}_1) & = \frac{1}{2} && \left( \sum_i \frac{\sigma_{0, i}^2 + 2 \sigma_{0, i}^2 v_i r_i}{\sigma_{1, i}^2} + \sum_i r_i^2 \sigma_i^2 \cdot \sum_i \frac{v_i^2}{\sigma_i^2} \right. \\
    & && + \sum_i \frac{(\bmu_1 - \bmu_0)_i^2}{\sigma_{1,i}^2} - k \notag \\
    & && + \left. 2 \left( \sum_i \left( \log \sigma_{1, i}^2 - \log \sigma_{0, i}^2 \right) - \log (1 + \sum_i v_i r_i) \right) \right) \notag.
\end{alignat}

\section{Supplementary to the experimental setup}
\label{sect:exp_details}

\subsection{Up-convolution}
We used convolutional inference networks for the cart-pole and three-link arm task. While these networks help us overcome the problem of large input dimensionalities (i.e. $2 \times 128 \times 128$ pixel images in the three-link arm task), we still have to generate full resolution images with the decoder network. For high-dimensional images generation fully connected neural networks are simply not an option. We thus decided to use up-convolutional networks, which were recently show to be powerful models for image generation \cite{Zeiler_CVPR2010,DosSprChair15,Kulkarni_arxiv2015}.

To set-up these models we basically ``mirror'' the convolutional architecture used for the encoder. More specifically for each $5 \times 5$ convolution followed by $2 \times 2$ max-pooling step in the encoder network, we introduce a $2 \times 2$ up-sampling and $5 \times 5$ convolution step in the decoder network. The complete network architecture is given below. It is similar to the up-convolution networks used in \citet{DosSprChair15}. The upsampling strategy we use is simple ``perforated'' upsampling as described in \cite{OseSoySma2014}.

\subsection{Variational Autoencoder with slowness}
Enforcing temporal slowness during learning has previously been found to be a good proxy for learning representations in reinforcement learning \cite{Jonschkowski14, Legenstein10}  and representation learning from videos \cite{NIPS2011Zou}.
We also consider a VAE variant with a slowness term on the latent representation by enforcing similarity of the encodings of temporally close images. This can be achieved by augmenting the standard VAE objective $\cL^{\text{bound}}$ with an additional KL divergence term on the latent posterior $Q_\bphi$:
\begin{align}
\cL^{\text{slow}}(\bx_t, \bx_{t+1}) = \text{KL}(Q_\bphi({\bz_{t+1}} | \bx_{t+1})  \| Q_\bphi({\bz_t} | \bx_t)).
\end{align}
Indeed there seems to be a slightly better coherence of similar states in the latent spaces, as e.g. depicted in Figure~\ref{fig:pole_latent_all} in the main paper. Yet, our experiments show that a slowness term alone does not suffice to structure the latent space, such that locally linear predictions and control become feasible.

\subsection{Evaluation criteria}
For comparing the performance of all variants of E2C and the baselines, the following criteria are of importance:
\begin{itemize}
    \item \textbf{Autoencoding}. Being able to reconstruct the given observations is the basic necessity for a model to work. The reconstruction cost drives a model to identify single states from its observations.
    \item \textbf{Decoding the next state}. For any planning to be possible at all, the decoder must be able to generate the correct images from transitions the dynamics model performed. If this is not the case, we know that the latent states of the encoding and the transition model do not coincide, thus preventing any planning.
    \item \textbf{Optimizing latent trajectory costs}. The action sequences for achieving a specified goal will be determined completely by locally linearized dynamics in the latent space. Therefore minimizing trajectory costs in latent space is, again, a necessity for successful control.
    \item \textbf{Optimizing real trajectory costs}. While the action sequence has been determined for the latent dynamics, the deciding criterion is whether this reflects the true state trajectory costs. Therefore carrying out the "dreamed" plans in reality is the optimality criterion for every model. To make the different models comparable, we use the same cost matrices for evaluation, which are not necessarily the same as for optimization.
\end{itemize}

We reflected these four criteria in the evaluation table in the paper. For the reconstruction of the current and next state we specified the mean log loss, which is in case of the Bernoulli distributions the cross entropy error function:
\begin{align}
    \log p(\bx | \hat{\bx}) = \frac{1}{N} \sum^N_{n=1} \sum^{n_x}_{i=0} x_{n, i} \log \hat{x}_{n, i} + (1 - x_{n, i}) \log (1 - \hat{x}_{n, i}).
\end{align}
For the costs a model imagines and truly achieves, we sample from different starting states and accumulate the distances in latent and true state space according to the SOC method.

\subsection{The three-link robot arm}
The robot arm we used in the last experiment in the main paper was
simulated using dynamics generated by the MapleSim
\url{http://www.maplesoft.com/products/maplesim/} simulator wrapped in
Python and visualized for producing inputs to E2C using PyGame.
We simulated a fairly standard robot arm with three links. The length
of the links were set to $2$, $1.2$ and $0.7$ (units were set to meters). The masses
of the corresponding links were all set to $10 kg$.

\subsection{Evaluating the true system model}
To compare the efficacy of different models when combined with optimal
control algorithms, we always reported the cost in latent space (as
used by the optimal control algorithm) as well as the ``real''
trajectory cost. To compute this real cost, we evaluated the same cost
function as in the latent space (quadratic costs on the deviation from
a given goal state), but using the real system states during execution and
different cost matrices for a fair comparison.

As an upper bound on the performance achievable for control by any
of the models, we also computed the true system cost by applying
iLQR/AICO to a model of the real system dynamics. We have this model
available since all experiments were performed in simulation.

\subsection{Neural Network training}
\subsubsection{Experimental Setup}
All the datasets were created in advance as $\mathcal{D} = \lbrace (\bx_1, \bu_1, \bx_{2}), \dots,  (\bx_{T-1}, \bu_{T-1}, \bx_{T}) \rbrace$ for the training, validation and test split. While the E2C models were trained on $\mathcal{D}$, the ones that do not incorporate any transition information (i.e. AE, VAE) were trained on images $\mathcal{D}_{\text{images}} = \lbrace \bx_1, \dots, \bx_{T} \rbrace$ extracted from the original dataset $\mathcal{D}$. The slowness VAE was trained on the pairs of images subset $\mathcal{D}_{\text{pairs}} = \lbrace (\bx_1, \bx_{2}), \dots, (\bx_{T-1}, \bx_{T}) \rbrace$ and our E2C models on the full $\mathcal{D}$.

In order to learn dynamics predictions for the image-only autoencoders, we extracted the latent representations and combined them with the actions from $\mathcal{D}$ into $\mathcal{D}_{\text{dynamics}} = \lbrace (\bz_1, \bu_1, \bz_{2}), \dots,  (\bz_{T-1}, \bu_{T-1}, \bz_{T}) \rbrace$. On these low-dimensional representations we trained the dynamics MLPs, thus ensuring that all methods were trained on exactly the same data.

\subsubsection{Implementation details}
We used orthogonal weight initialization for every layer \cite{Saxe13}. As described in the main paper, Adam \cite{Kingma_2015} was used as the learning rule for all networks. We found both these techniques to be fundamentally important for stabilizing training and achieving good reconstructions for all methods. Both methods also clearly helped to cut the hyperparameter search needed for all methods to a minimum. In the process of training, we could make out three phases: the unfolding of the latent space, the overcoming of the trivial solution (the average image of the dataset) and the minimization of the latent KL term. The architectures used for our experiments were as follows (where ReLU stands for rectified linear units \cite{GlorotBB11} and conv. for convolutions):

\paragraph{Plane}
\begin{itemize}[noitemsep]
	\item Input: $40^2$ image dimensions, $2$ action dimensions
    \item Latent Space dimensionality: $2$
	\item Encoder: 150 ReLU - 150 ReLU - 150 ReLU - 4 Linear (2 for AE)
	\item Decoder: 200 ReLU - 200 ReLU - 1600 Linear (Sigmoid for AE)
	\item Dynamics: 100 ReLU - 100 ReLU + Output layer (except Global E2C)
		\begin{itemize}[noitemsep] 
			\item AE, VAE, VAE with slowness, Non-linear E2C: 2 Linear
			\item E2C: 8 Linear ($2 \cdot 2$ for $\bA_t$, $2
                          \cdot 1$ for $\bB_t$, 2 for $\bo_t$), $\lambda = 0.25$
		\end{itemize}
	\item Adam: $\alpha = 10^{-4}, \beta_2=0.1$
    \item Evaluation costs: $\bR_z = 0.1 \cdot \bI$, $\bR_u = \bI$, $\bR_o = \bI$
\end{itemize}

\paragraph{Pendulum swing-up}
\begin{itemize}[noitemsep]
	\item Input: $2 \cdot 48^2$ image dimensions, $1$ action dimension
    \item Latent Space dimensionality: $3$
	\item Encoder: 800 ReLU - 800 ReLU - 6 Linear (3 for AE)
	\item Decoder: 800 ReLU - 800 ReLU - 4608 Linear (Sigmoid for AE)
	\item Dynamics: 100 ReLU - 100 ReLU + Output layer (except Global E2C)
		\begin{itemize}[noitemsep] 
			\item AE, VAE, VAE with slowness, Non-linear E2C: 3 Linear
			\item E2C: 12 Linear ($2 \cdot 3$ for $\bA_t =
                          (\bI + \bv_t \br_t^T)$, $3 \cdot 1$ for
                          $\bB_t$, 3 for $\bb_t$), $\lambda = 0.25$
		\end{itemize}
	\item Adam: $\alpha = 3 \cdot 10^{-4}, \beta_2=0.1$
    \item Evaluation costs: $\bR_z = \bI$, $\bR_u = 0.1 \bI$
\end{itemize}

\paragraph{Cart-Pole balancing}
\begin{itemize}[noitemsep] 
	\item Input: $2 \cdot 80^2$ image dimensions, $1$ action dimension
    \item Latent Space dimensionality: $8$
	\item Encoder: $32 \times 5\times 5$ ReLU -  $32 \times 5 \times 5$ ReLU - $32 \times 5 \times 5$ ReLU - 512 ReLU - 512 ReLU
	\item Decoder: 512 ReLU - 512 ReLU - $2 \times 2$ up-sampling - $32 \times 5 \times 5$ ReLU - $2 \times 2$ up-sampling - $32 \times 5 \times 5$ ReLU - $2 \times 2$ up-sampling - $32 \times 5\times 5$ conv. ReLU
	\item Dynamics: 200 ReLU - 200 ReLU + 32 Linear ($2 \cdot 8$ for $\bA_t = (\bI + \bv_t \br_t^T)$, $8 \cdot 1$ for $\bB_t$, $8$ for $\bb_t$), $\lambda = 1$
	\item Adam: $\alpha = 10^{-4}, \beta_2=0.1$
    \item Evaluation costs: $\bR_z = \bI$, $\bR_u = \bI$
\end{itemize}

\paragraph{Three-link arm}
\begin{itemize}[noitemsep] 
	\item Input: $2 \cdot 128^2$ image dimensions, $3$ action dimensions
        \item Latent Space dimensionality: $8$
	\item Encoder: $64 \times 5\times 5$ conv. ReLU - $2 \times 2$ max-pooling -  $32 \times 5 \times 5$ conv. ReLU - $2 \times 2$ max-pooling - $32 \times 5 \times 5$ conv. ReLU - $2 \times 2$ max-pooling - 512 ReLU - 512 ReLU
	\item Decoder:  512 ReLU - 512 ReLU - $2 \times 2$ up-sampling - $32 \times 5 \times 5$ ReLU - $2 \times 2$ up-sampling - $32 \times 5 \times 5$ ReLU - $2 \times 2$ up-sampling - $64 \times 5\times 5$ conv. ReLU
	\item Dynamics: 200 ReLU - 200 ReLU + 48 Linear ($2 \cdot 8$ for $\bA_t = (\bI + \bv_t \br_t^T)$, $8 \cdot 3$ for $\bB_t$, $8$ for $\bb_t$), $\lambda = 1$
	\item Adam: $\alpha = 10^{-4}, \beta_2=0.1$
    \item Evaluation costs: $\bR_z = \bI$, $\bR_u = 0.001\bI$
\end{itemize}

\begin{figure}[H]
	\centering
	\includegraphics[scale=0.7]{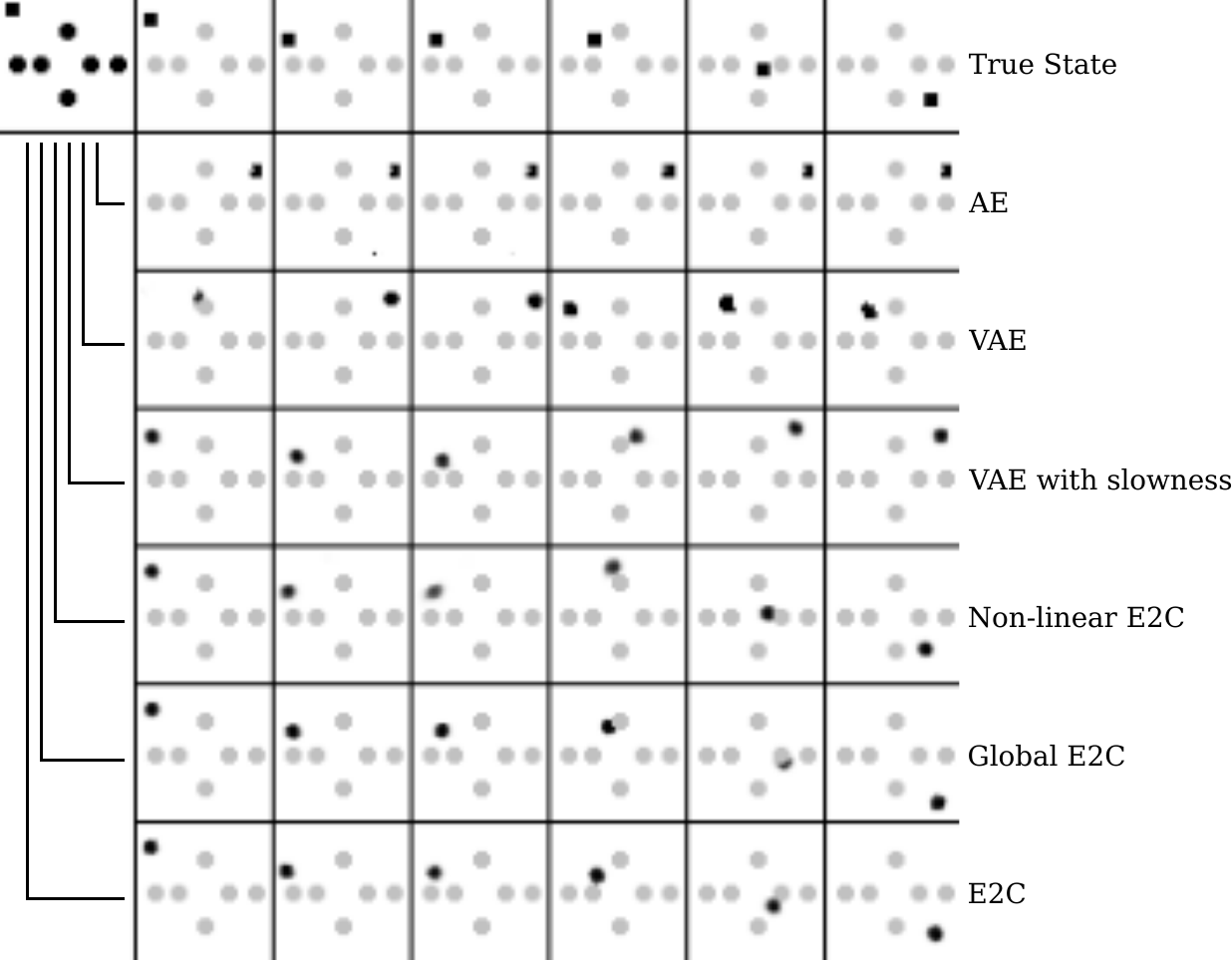}
	\caption{Generated ``dreamed'' trajectories of different models for the plane task (from left to right). The opacity of the obstacles has been lowered in this depiction for better visibility of the agent.}
	\label{fig:plane_trajectories}
\end{figure}

\section{Supplementary evaluations}
\label{sect:eval}
\subsection{Trajectories for plane and pendulum}
\label{sect:traj}
\begin{figure}[H]
	\centering
	\includegraphics[scale=0.3]{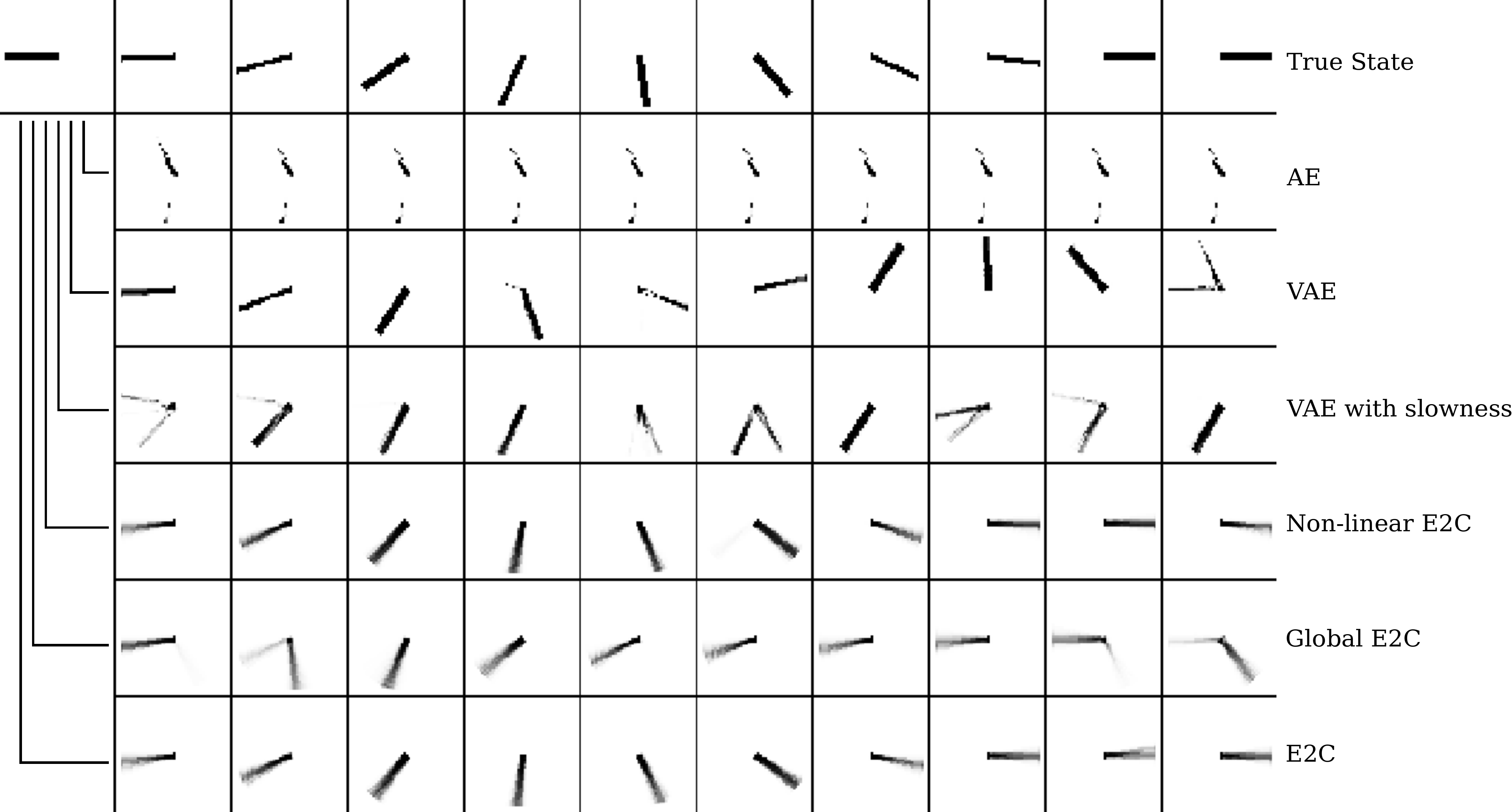}
	\caption{Generated ``dreamed'' trajectories (from left to right) for \emph{passive} dynamics: the pendulum starts with angle $\theta = -\frac{\pi}{2}$ without velocity. The models have to predict the dynamics, while no force is applied.}
	\label{fig:pole_traj_passive}
\end{figure}

\begin{figure}[H]
	\centering
	\includegraphics[scale=0.3]{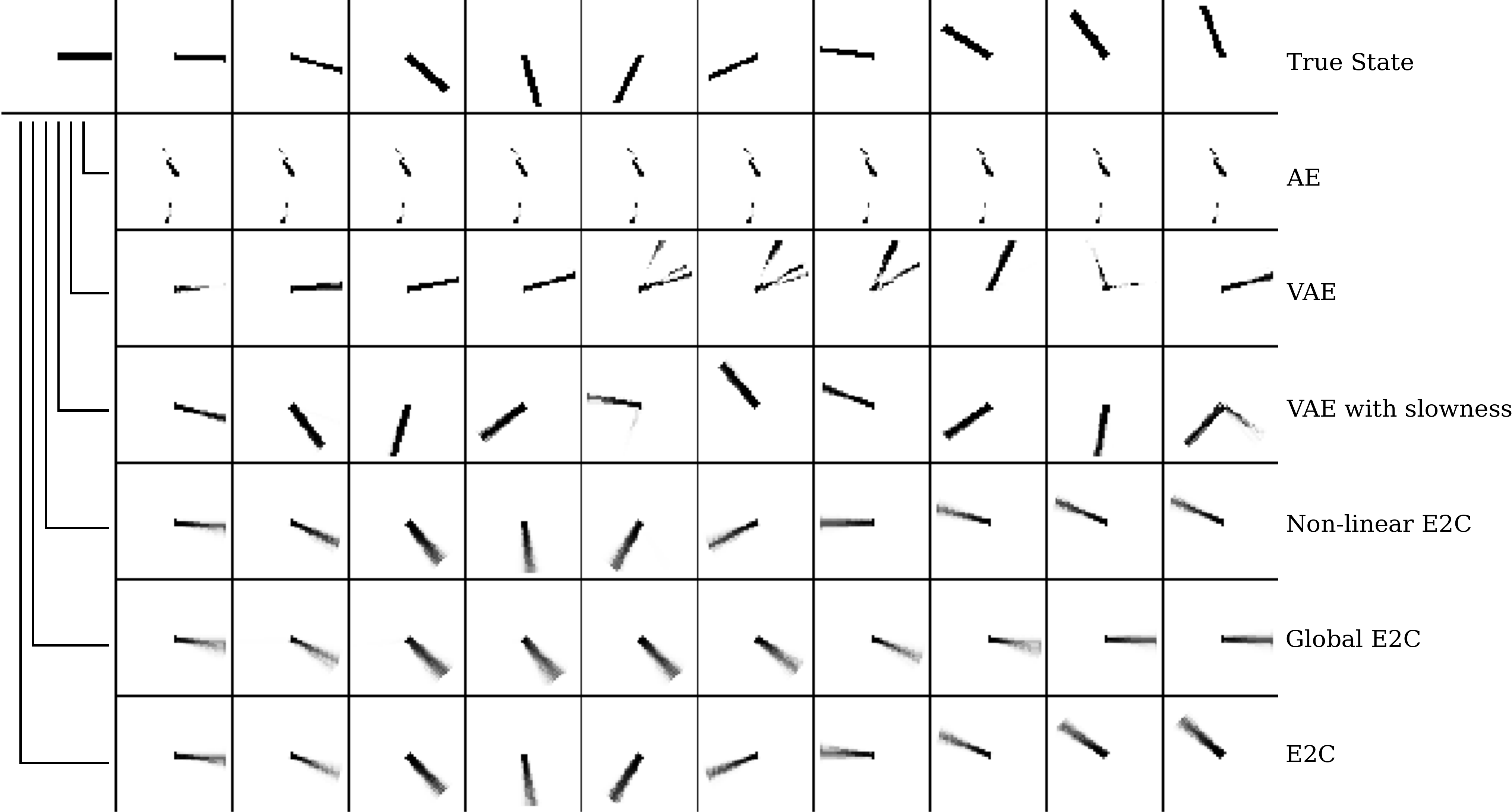}
	\caption{Dreamed trajectories (from left to right) for \emph{controlled} dynamics: the pendulum starts with angle $\theta = \frac{\pi}{2}$ without velocity. For 6 timesteps, full force is applied to the right, followed by 4 timesteps of full force to the left.}
	\label{fig:pole_traj_active}
\end{figure}

To qualitatively measure the predictive accuracy, the starting state for a trajectory is encoded and the actions are applied on the latent representation. After each transition, the predicted latent position is decoded and visualized. In this manner, multi-step predictions can be generated for the planar system in Figure \ref{fig:plane_trajectories} and for the inverted pendulum in Figures \ref{fig:pole_traj_passive} and \ref{fig:pole_traj_active}.

\subsection{Inverted pendulum latent space}
\begin{figure}[h]
	\centering
	\includegraphics[scale=0.32]{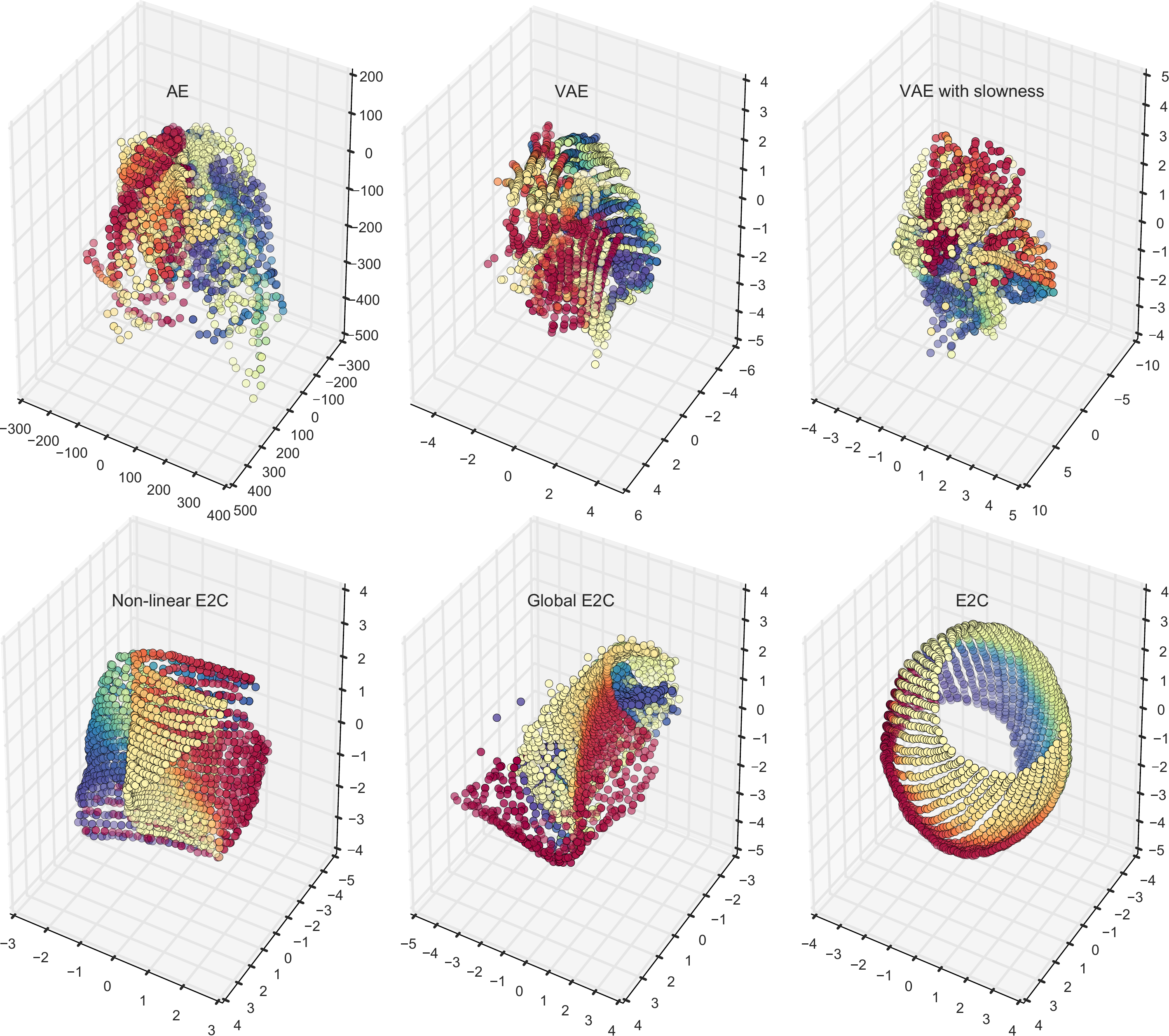}
	\caption{Latent spaces of all baseline models and E2C variants for the inverted pendulum.}
        \label{fig:pole_latent_all}
\end{figure}

Encoding the pendulum depictions into a 3-dimensional latent space allows for
a visual comparison in Figure \ref{fig:pole_latent_all}~.
\label{sect:vis_latent}

\subsection{Trajectories for cart-pole and three-link arm}
\begin{figure}
	\centering
	\includegraphics[scale=0.3]{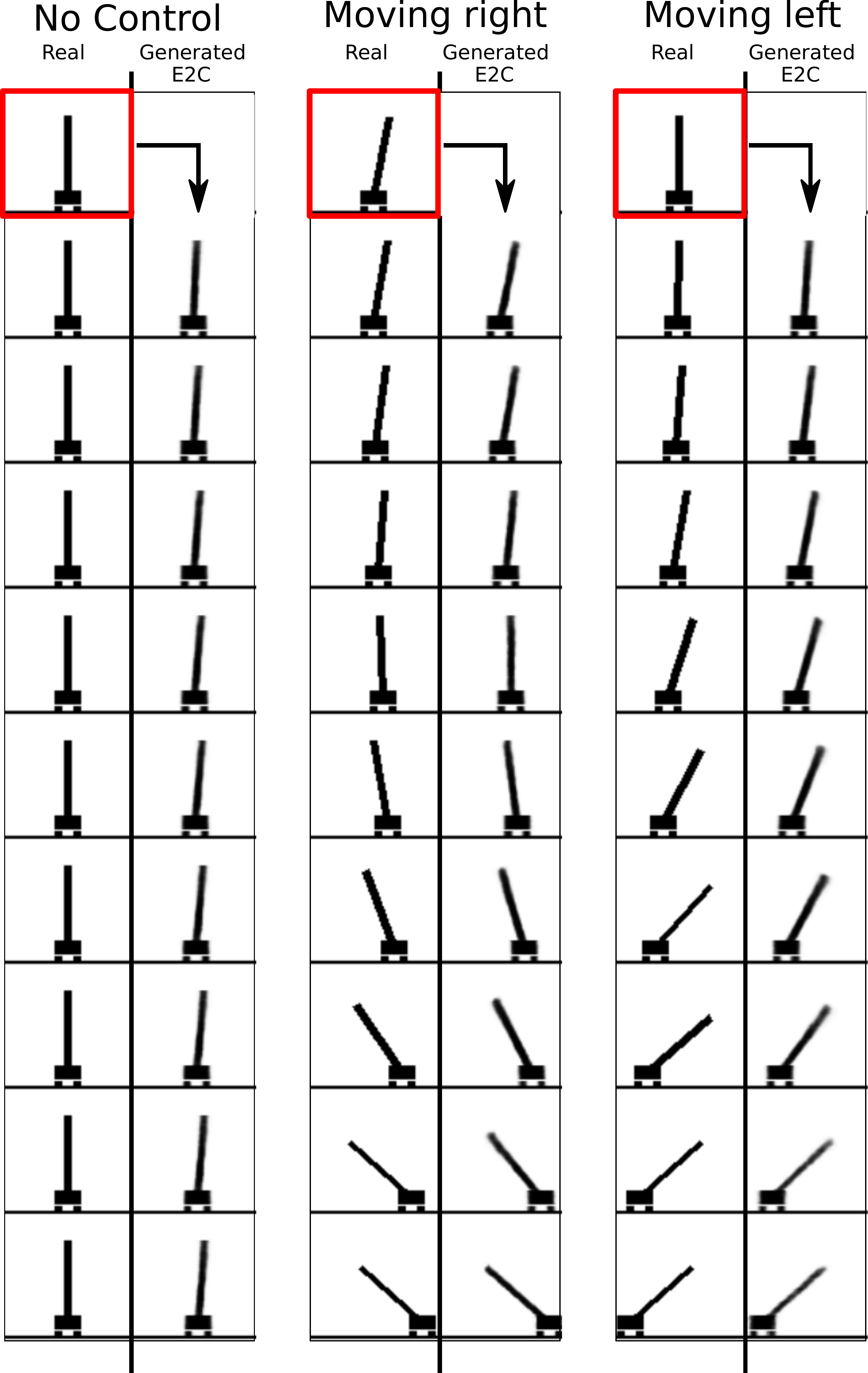}
	\caption{Dreamed trajectories (top to bottom) for uncontrolled (left column) and
          controlled (middle/right column) dynamics in the cart-pole system. The red image shows the initial configuration, which is encoded resulting in $\bz_1$. The images in the right half of each column are then generated without additional input by following the dynamics in latent space. The left column depicts the uncontrolled case ($\bu = 0$ for all steps). The middle column shows a controlled trajectory with torque $-20$ applied in each step and the right column a trajectory with torque $20$ applied in each step. Prediction of the history image is omitted in these depictions. }
	\label{fig:cpole_traj}
\end{figure}

Finally -- similar to the images in Section \ref{sect:traj} -- Figure \ref{fig:cpole_traj} shows multi-step predictions for the cart-pole system. We depict important cases: (1) a long-term prediction with the cart-pole standing still (essentially the unstable fix-point of the underlying dynamics); (2) the cart-pole moving to the right, changing the direction of the poles angular velocity (middle column); (3) and the pole moving farthest to the right. The long-term predictions by the E2C model are all of high quality. Note that for the uncontrolled dynamics the predictions show a slight bias of the pole moving to the right (an effect that we consistently saw in trained models for the cart-pole). We attribute this problem to the fact that discretization errors in the image rendering process of the pole angle make it hard to predict small velocities accurately.

\subsection{Exemplary trajectory taken for three-link arm task}
\begin{figure}[H]
	\centering
	\includegraphics[scale=0.5]{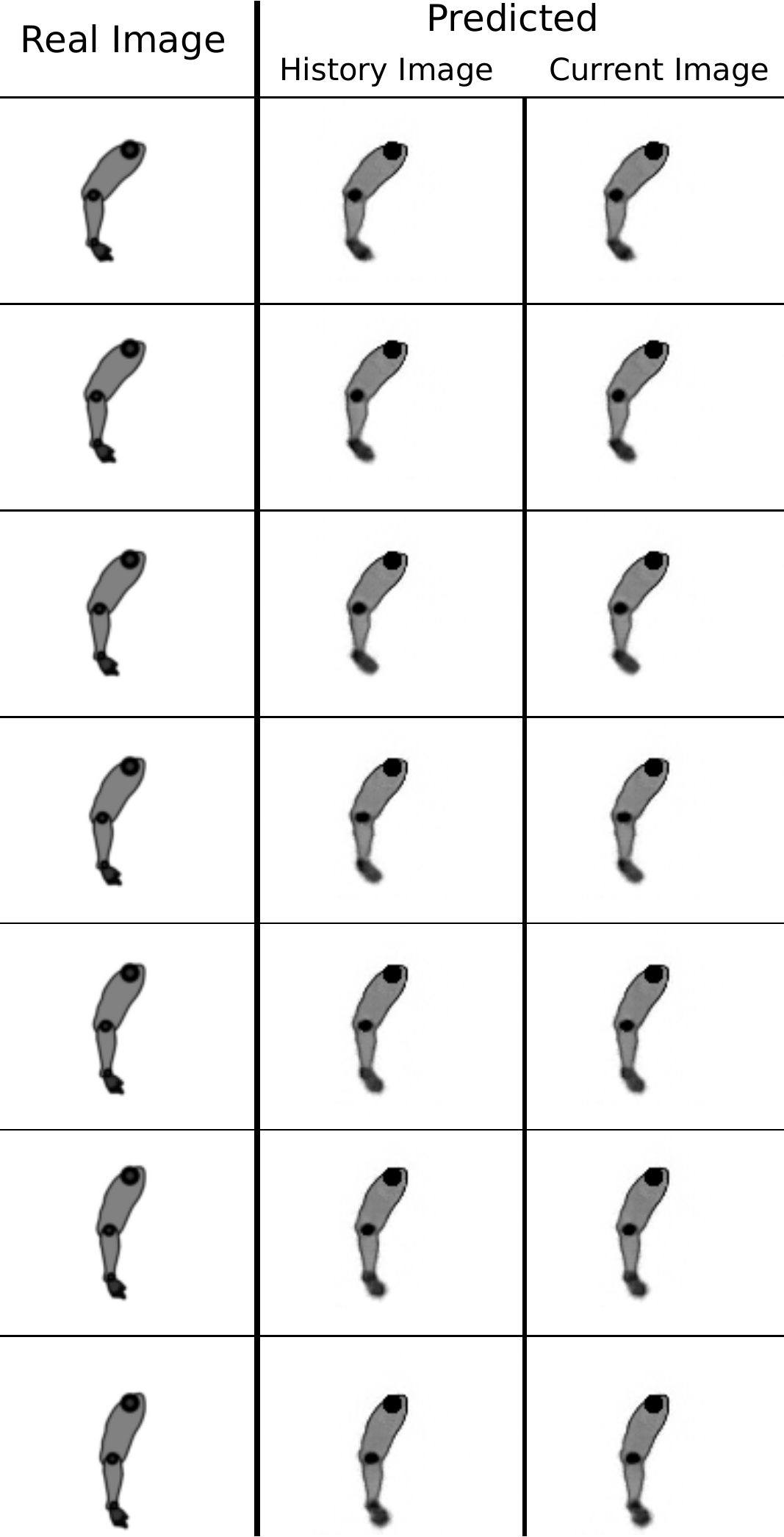}
        \caption{Frames extracted from a trajectory (top to bottom) as executed by the
        Embed to Control system. The left column shows the real images
      corresponding to transitions taken in the MDP. Middle and right
      column show the prediction of history image and current image
      based on the previous two images.}
        \label{fig:arm_traj}
\end{figure}

Figure \ref{fig:arm_traj} shows a segment of a controlled trajectory for the three-link arm as executed by the E2C system. Note that, in contrast to other figures in this supplementary material, it does \textbf{not} show a long-term prediction but rather 10 steps of a trajectory (together with one-step-ahead predictions) that was taken by the E2C system when combined with model predictive control. For additional visualizations and controlled trajectories for all tasks we refer to the supplementary video.

\subsection{Comparison of different models for cart-pole and robot
  arm}
\label{sec:models_comp_cp_arm}
In Table \ref{tab:traj_costs} we compare our variety of models 
in terms of real trajectory cost and task success percentage on the cart-pole
and the robot arm.
All results are averaged over 30 different starting states with a fixed goal state.

The cart-pole 
always starts in the goal state (zero angle and zero velocity) with small
additive Gaussian noise ($\sigma = 0.01$). Success is defined as preventing the pole from 
falling below an angle of $\pm 0.85$ rad.
The three-link arm system begins in  a random configuration and the goal is to to unroll all 
joints (e.g. make all angles zero) and stay $\epsilon$-close to that position.

The results show that only E2C and its 
non-linear variant can perform this task successfully, although there is still a 
large performance gap between the two. We conclude, that the error of 
linearizing non-linear dynamics after training the corresponding model grows 
to the point of no longer allowing accurate control for the system.

\begin{table}
\begin{center}
\caption{Comparison between trajectory costs of different approaches for the cart-pole and three-link task. The standard Autoencoder, Variational Autoencoder and Global E2C model are omitted from the table as they failed on this task (performance similar to VAE with slowness).}
\label{tab:traj_costs}
\vspace{0.1cm}
\footnotesize
\begin{tabular}{l|c|cccc}
\textbf{Algorithm} & True model & VAE + slownes & E2C no latent KL & Non-linear E2C & E2C \\
\hline
& & \multicolumn{4}{c}{\bf Cart-Pole balance}\\
\textbf{Traj. Cost} & 15.33 $\pm$ 7.70 & 49.12 $\pm$ 16.94 & 48.90 $\pm$ 17.88 & 31.96 $\pm$ 13.26 & \textbf{22.23 $\pm$ 14.89} \\
\textbf{Success $\%$} & 100 $\%$ & 0 $\%$ & 0 $\%$ & 63 $\%$ & \textbf{93 $\%$} \\
\hline
& & \multicolumn{4}{c}{\bf Three-link arm}\\
\textbf{Traj. Cost} & 59.46 & 1275.53 $\pm$ 864.66 & 1246.69 $\pm$ 262.6  & 460.40 $\pm$ 82.18  & \textbf{90.23 $\pm$ 47.38} \\
\textbf{Success $\%$} & 100 $\%$ & 0 $\%$ & 0 $\%$ & 40 $\%$ & \textbf{90 $\%$} \\
\end{tabular}
\end{center}
\end{table}

\subsection{Comparison of trajectory optimizers for cart-pole and robot arm}
To compare how well AICO deals with the covariance matrices estimated
in latent space we performed an additional experiment on the
cart-pole and three-link robot arm task comparing it to iLQR. We
performed model predictive control using the locally linear E2C model
starting in 10 different start states each. The remaining settings are as given
in Section \ref{sec:models_comp_cp_arm}.

As reported in Table \ref{tab:aico_ilqr}, both methods performed about
the same for these tasks, indicating that the covariance matrices estimated by our model do not ``hurt'' planning, but considering them does not improve performance either.

\begin{table}
\begin{center}
\caption{Comparison between AICO and iLQR based on the ``real'' cost for controlling the
  cart-pole and three-link robot arm using convolutional networks.}
\label{tab:aico_ilqr}
\begin{tabular}{lcc}
 \multicolumn{1}{c}{Method} & \multicolumn{1}{c}{iLQR} & \multicolumn{1}{c}{AICO}
  \\
\hline
  \multicolumn{3}{c}{\textbf{Cart-Pole}} \\
\hline
 E2C & $14.56 \pm 4.12$ & $12.56 \pm 2.47$ \\
 True model & $7.45 \pm 1.22$ &  $7.03 \pm 1.07$ \\
\hline
  \multicolumn{3}{c}{\textbf{Three-Link Robot Arm}} \\
\hline
 E2C & $93.78 \pm 32.98$ & $92.99 \pm 20.12$ \\
 True model & $53.59 \pm 9.74$ & $56.34 \pm 10.82$ \\
\end{tabular}
\end{center}
\end{table}

\end{appendix}

\end{document}